\documentclass[10pt,twocolumn,letterpaper]{article}

\usepackage[pagenumbers]{cvpr} 

\usepackage{graphicx}
\usepackage{booktabs}
\usepackage{xcolor,colortbl}
\usepackage{comment}
\usepackage{amsmath}
\usepackage{amssymb}
\usepackage{booktabs}
\usepackage{multirow}
\usepackage{tabularx}
\usepackage{bm}
\usepackage{wrapfig}
\usepackage{physics}
\usepackage{breqn}

\definecolor{blue}{rgb}{0.21,0.49,0.74}
\definecolor{lightgray}{rgb}{0.88, 0.88, 0.88}
\definecolor{yellow2}{rgb}{1, 0.925, 0.6}
\usepackage[pagebackref,breaklinks,colorlinks,allcolors=blue]{hyperref}

\def\paperID{****} 
\def\confName{Arxiv}
\def\confYear{2025}

\newcommand*{\method}{EgoH4\xspace}

\title{The Invisible EgoHand: 3D Hand Forecasting through EgoBody Pose Estimation}

\author{
Masashi Hatano\textsuperscript{1}\thanks{Work done during a research visit to the University of Bristol}
\quad
Zhifan Zhu\textsuperscript{2}
\quad
Hideo Saito\textsuperscript{1}
\quad
Dima Damen\textsuperscript{2}
\\
\textsuperscript{1} Keio University \quad
\textsuperscript{2} University of Bristol
\\
\url{https://masashi-hatano.github.io/EgoH4}
}

\begin{document}
\maketitle
\begin{abstract}
Forecasting hand motion and pose from an egocentric perspective is essential for understanding human intention.
However, existing methods focus solely on predicting positions without considering articulation, and only when the hands are visible in the field of view.
This limitation overlooks the fact that approximate hand positions can still be inferred even when they are outside the camera's view. 
In this paper, we propose a method to forecast the 3D trajectories and poses of both hands from an egocentric video, 
both in and out of the field of view.

We propose a diffusion-based transformer architecture for Egocentric Hand Forecasting, \method, which takes as input the observation sequence and camera poses, then predicts future 3D motion and poses for both hands of the camera wearer.
We leverage full-body pose information, allowing other joints to provide constraints on hand motion.
We denoise the hand and body joints along with a visibility predictor for hand joints and a 3D-to-2D reprojection loss that minimizes the error when hands are in-view.

We evaluate \method on the Ego-Exo4D dataset, combining subsets with body and hand annotations.
We train on 156K sequences and evaluate on 34K sequences, respectively.
\method improves the performance by 3.4cm and 5.1cm over the baseline in terms of ADE for hand trajectory forecasting and MPJPE for hand pose forecasting.
\end{abstract}    
\begin{figure}[t]
    \centering
    \includegraphics[width=\linewidth]{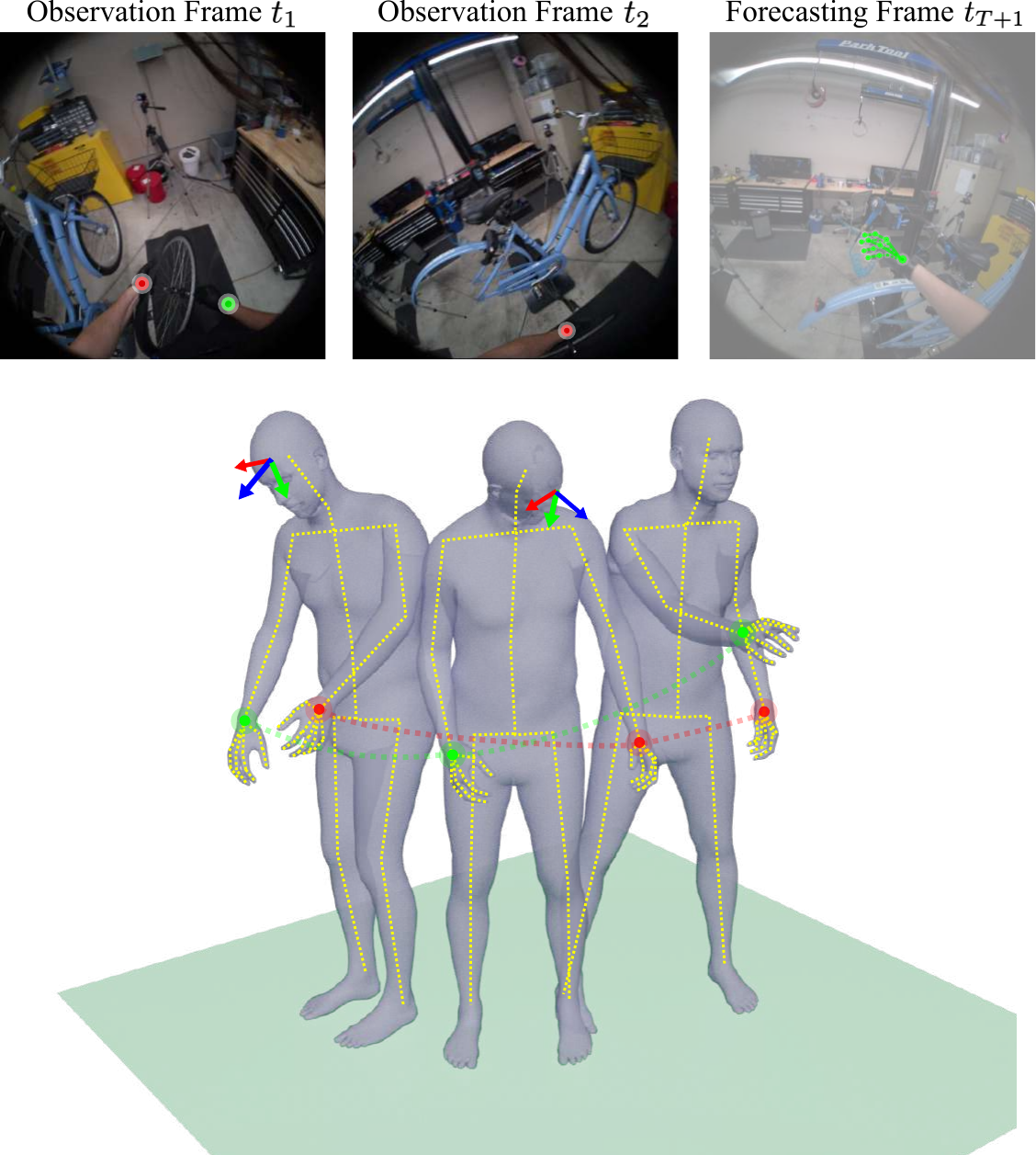}
    \caption{
    Given signals during observation: camera poses, images, and visible hand locations in 2D, 
    our proposed method \textbf{\method} forecasts future 3D hand pose.
    \method can forecast hand joints even when hands are out of view during observation.
    We show visible 2D hand positions overlaid on the observation frames $t_1$ and $t_2$, and the corresponding camera poses attached on the heads.
    At $t_2$, the right hand is invisible.
    In the forecasting frame, the right hand is back in view while the left hand is now out of view.
    }
    \vspace*{-12pt}
    \label{fig:teaser}
\end{figure}

\section{Introduction}
\label{sec:intro}
Understanding human motion from an egocentric video is critical for a variety of applications, including AR/VR, human-robot interaction, and assistive technology.
Unlike category-level discrete predictions, such as action recognition~\cite{hamed2012cvpr, poleg2016wacv, Plizzari2023iccv, tekin2019cvpr, hatano2024eccv} or action anticipation~\cite{roy2024interaction, thakur2024leveraging, furnari2019would, girdhar2021anticipative, wu2020learning}, motion provides fine-grained, continuous predictions.

Prior works predicting future hand positions are 2D-based, estimating the location within a moving camera frame. 
These works focus on hand-object interactions~\cite{liu2020forecasting, liu2022joint} or ego-motion awareness~\cite{hatano2024eccvw, ma2024diff, ma2024madiff}.
However, 2D methods are unable to give persistent predictions of the hand in the environment due to the camera motion.
Instead, recent works have targeted predicting in a world coordinate frame, with successful methods targeting body pose~\cite{khirodkar2023ego, mueller2023buddi, goel2023humans, ye2023slahmr, rempe2021humor},
ego body pose~\cite{hakada2022unrealego, wang2021estimating, yuan20183d, cuevas2024simpleego, zhang2023probabilistic, li2023ego, castillo2023bodiffusion} and recently 3D hand trajectories~\cite{BaoUSST_ICCV23}.
Works that focus on 3D forecasting body pose~\cite{yuan2019ego, choudhury2023tempo, escobar2025egocast}, hand location or pose~\cite{BaoUSST_ICCV23, tang2025prompting} remain scarce.

The current egocentric hand forecasting task has three significant limitations we address here:

\noindent \textbf{Out-of-view}.
Previous works~\cite{BaoUSST_ICCV23, liu2022joint, liu2020forecasting} have not considered scenarios where hands move out of the field of view during observations, yet this is a common occurrence in egocentric videos (as shown in \cref{fig:teaser}).
Hands are visible primarily when interacting with objects or just before intentional contact.
For example, when reaching for distant objects, we initially move our body closer, only extending our hand once we are within reach.
Consequently, relying on visible observations for forecasting can result in delayed predictions, limiting early-stage forecasting capabilities.

\noindent \textbf{Body Movements Awareness}.
Given the natural coordination between the hands and body, incorporating body movements could enhance the accuracy of hand forecasting.
Predicting body poses alongside hand poses helps prevent unrealistic hand motions as body joints serve as the constraints.

\noindent \textbf{Hand Articulation}.
Prior egocentric hand forecasting works~\cite{BaoUSST_ICCV23, liu2022joint, liu2020forecasting} predict a hand position without finger articulation.
While HoloAssist~\cite{HoloAssist2023} recently established a benchmark for egocentric hand pose forecasting without introducing a specific method, the task of hand pose forecasting in egocentric videos is still largely underexplored.

To address these limitations, we propose \textbf{\method}, a 3D hand forecasting method that leverages body pose estimation to predict 3D hand motion and pose.
To our knowledge, this is the first work to attempt 3D hand trajectory and pose forecasting when hands are out of the field of view, i.e., \textit{invisible}.
We achieve this by (1)~jointly optimizing the hands and body joints. Leveraging the body pose knowledge helps locate the hand joints when they are out of frame by constraining hand joints relative to the body pose.
Additionally, we (2) incorporate a classifier that estimates hand visibility.
This improves the capability of dealing with invisible hands and enhancing hand forecasting accuracy.

We evaluate \method on the Ego-Exo4D~\cite{Grauman_2024_CVPR} dataset, which offers 3D annotations even when hands are outside the camera's field of view thanks to the multiple exocentric cameras.
In summary, our contributions are:
\begin{itemize}
    \item We are the first to address egocentric 3D hand forecasting when hands are in- or out-of-view, during both the observation and the forecasting timesteps.
    \item We also extend the task of egocentric hand trajectory forecasting to hand pose forecasting for a fine-grained understanding of human intention.
    \item We propose \method, a diffusion-based transformer model jointly denoising body pose and hands along with a visibility predictor and 3D-to-2D reprojection regularization.
    \item We evaluate \method on the Ego-Exo4D dataset, a large-scale egocentric dataset. From available annotations, we curate a 3D hand forecasting task, resulting in 156K training sequences and 34K testing sequences.
    \item We improved the hand trajectory forecasting accuracy in ADE by 5.5cm, 1.9cm, and 3.4cm, and hand pose forecasting in MPJPE by 5.0 cm, 5.9 cm, and 5.1 cm for in-view, out-of-view, and overall sequences, respectively.
\end{itemize}

\section{Related Work}
\label{sec:related}

We review related works on hands in egocentric videos, egocentric hand forecasting, motion forecasting, and ego-body pose estimation.

\subsection{Understanding Hands in Egocentric Videos}
Hand-object interactions are best studied in egocentric videos.
Prior works have addressed 2D hand detection and side classification~\cite{shan20_100doh, cheng23_handv2, mmpose2020},
hand segmentation~\cite{VISOR2022, zhang2022fine, jia2022generative}, grasp type classification~\cite{goyal2022human} and hand pose estimation~\cite{pavlakos2024reconstructing, rong2021frankmocap, prakash2024hands, oh2023BlurHand, Park_2022_CVPR_HandOccNet}. 
Other works also involve modeling hand-object interactions~\cite{goletto2024amego, nagarajan2023egoenv, fan2025benchmarks, sener2022assembly101, doosti_2020_cvpr, FirstPersonAction_CVPR2018} and object affordance~\cite{goyal2022human, ye2023affordance}.
These works provide robust methods to solve 2D hand-related tasks for egocentric videos.

\subsection{Egocentric Hand Forecasting}
Initial efforts~\cite{liu2020forecasting,liu2022joint} in 2D hand trajectory forecasting from egocentric videos aim to understand human intention, often combined with interaction hotspot prediction and action anticipation, as hand motion is a key cue for anticipation. 
Recent works~\cite{ma2024diff, ma2024madiff, hatano2024eccvw} consider the ego-motion, the head motion of the human, to improve the 2D hand trajectory.
However, predicting hand locations only in the 2D image plane limits the range of forecasting, as hands move in a much wider range in 3D space.

USST~\cite{BaoUSST_ICCV23}  is the first work to address 3D hand forecasting and propose a pipeline to lift up from a manually annotated 2D hand landmark into 3D to acquire training data.
Also, they propose the uncertainty-aware state space transformer model that takes the 3D hand trajectory and egocentric videos as input and forecasts the 3D hand trajectory.
However, their scope is limited to position without pose and only when the hands are visible.

We are the first to address 3D hand forecasting even when hands are out-of-view during observation, and the first to explore forecasting hand poses (not only positions).

\begin{figure*}[t]
\centering
   \includegraphics[width=0.9\linewidth]{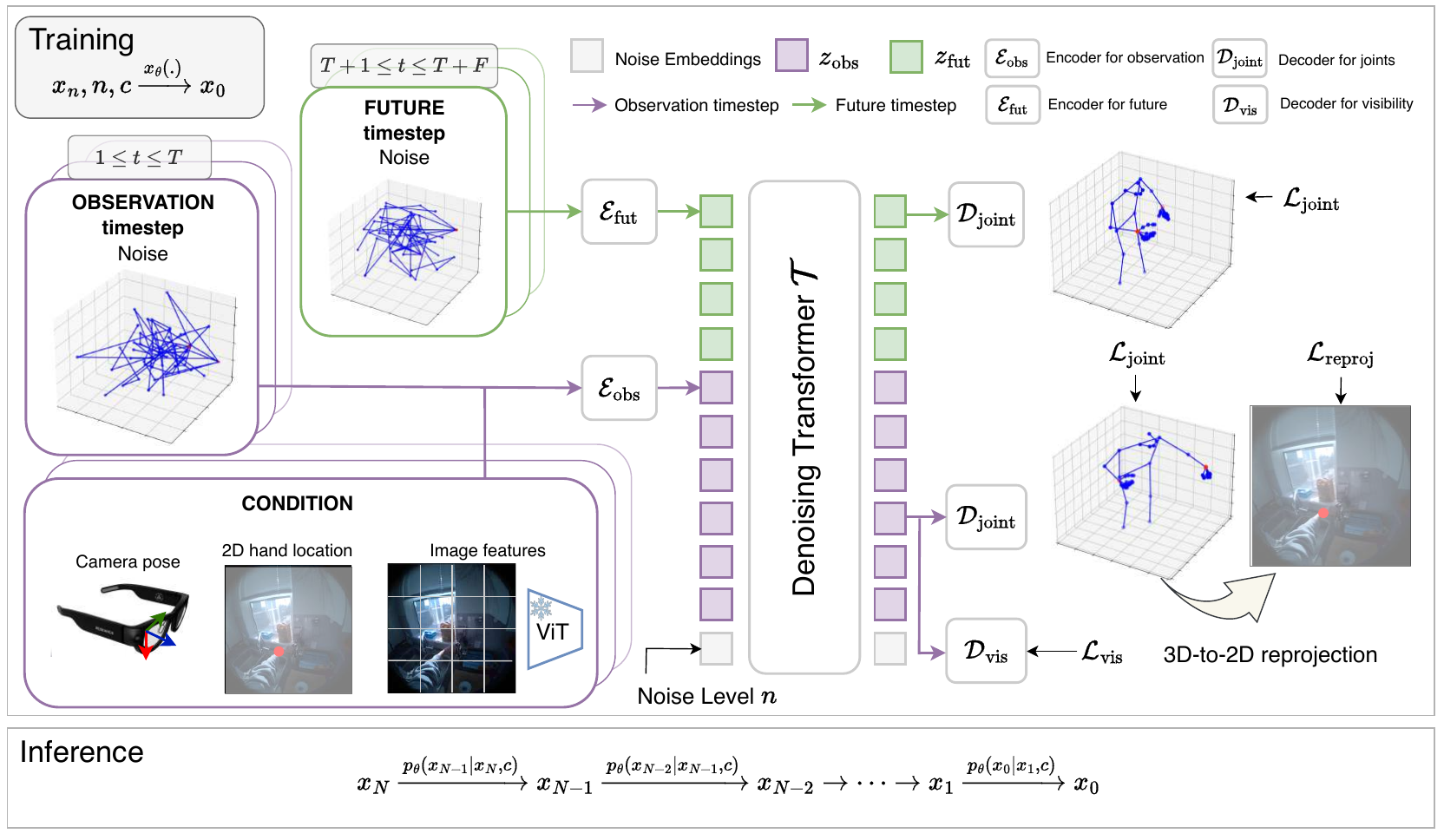}
\vspace*{-6pt}
   \caption{\textbf{The framework of our proposed method, \method.} We show the denoising network in a single denoising step. During training, we estimate the original data $x_0$ from an arbitrary noise level $n$ to learn the denoising network. During inference, we iteratively denoise the noisy joints over the maximum diffusion step $N$ from $N$ to $0$.}
   \vspace*{-12pt}
\label{fig:overview}
\end{figure*}

\subsection{Body Motion Forecasting}
Forecasting human motion has gained significant attention from various perspectives~\cite{chen2023humanmac, diller2023futurehuman3d, mao2021generating, hu2024hoimotion}. While most studies focus on body pose, a few address hand articulation~\cite{tang2025prompting} or integrate it with body pose~\cite{yan2024forecasting}. Among these, GCN-based methods~\cite{li2022skeleton, ma2022progressively} are widely used and effective, where joints serve as nodes and edges capture spatial relationships.
Other works~\cite{cao2020long, yang20223d} first predict 2D keypoints from images and estimate 3D poses, subsequently forecasting future motions based on the estimated past motions.
However, these methods make strong assumptions: past motions are accessible or body joints are visible, limiting their applicability in egocentric scenarios.

On the other hand, several works~\cite{yuan2019ego, escobar2025egocast} address body pose forecasting from an egocentric perspective.
EgoCast~\cite{escobar2025egocast} proposes a two-stage approach: first estimate the body pose of the camera wearer, and then forecast motion from the estimated body poses.
These works~\cite{yuan2019ego, escobar2025egocast} do not predict hand articulation or report performance on hand position specifically.

\subsection{Egocentric 3D Body Pose Estimation}
There has been a growing interest in 3D human body pose estimation from egocentric videos.
Several works~\cite{hakada2022unrealego, wang2021estimating, wang2024egocentric} utilize the fisheye camera, which provides a large field of view to ensure body joints are visible.
EgoPoser~\cite{jiang2024egoposer} uses a wearable device with hand joint locations as input.
You2Me~\cite{ng2020you2me} focuses on estimation via interaction with other people~\cite{Zhang2022EgoBody}.
Other works use deep RL-based approach~\cite{yuan2019ego} or physically plausible predictions~\cite{Luo2021DynamicsRegulatedKP}.

EgoEgo~\cite{li2023ego} is the first work that proposed a diffusion-based body generation model conditioned on head poses. 
Other works~\cite{castillo2023bodiffusion, jiang2024egoposer, yi2024egoallo, chi2025estimating} investigate this task using ground-truth head pose as a condition.
This can be provided by smart glasses devices like Project Aria~\cite{engel2023projectarianewtool}.

In this work, we also estimate the egocentric human body poses from head poses, and we leverage recent advancement in egocentric body pose estimation using diffusion models; however, our focus is not on improving the body estimation but leveraging it for 3D hand forecasting.

\newcommand{\condOBS}{\mathcal{C}_{obs}}
\newenvironment{OldPart}{\leavevmode\color{gray}\ignorespaces}{}

\section{Our Method, \method}
\label{sec:method}
We propose \method, a diffusion-based transformer architecture for 3D hand forecasting, which takes head poses, 2D hand locations, and image features as input during the observation period and aims to predict future 3D hand poses. 

We first introduce a novel egocentric 3D hand pose forecasting problem setup (\cref{sec:problem-def}).
Then, we introduce our proposed method, \method, with full body motion estimation~(\cref{sec:cdm}), and training objective (\cref{sec:objective-fn}). 
\cref{fig:overview} provides an overview of our approach. 

\subsection{Problem Setup}
\label{sec:problem-def}
Given an input egocentric video and the corresponding camera poses for $T$ observation frames, the goal is to forecast the 3D hand poses $Y_{\text{fut}}=\{y^{T+1}, \dots, y^{T+F}\}$ for the future time horizons $F$.
At each timestep $T+1 \le t \le T+F$, $y^{t}$ consists of left/right hand joints.
Importantly, we employ sequence canonicalization~\cite{li2023ego, yi2024egoallo}, where 3D points are expressed relative to a canonicalized coordinate system from the first camera coordinate.

\subsection{\method Architecture}
\label{sec:cdm}
Leveraging knowledge of 3D body movements helps enhance hand forecasting when the hand is out of view, during observation, or forecasting.
Since body joints are mostly invisible and hand visibility frequently changes, there is no direct, deterministic mapping function for estimating these joints from these observations.
Therefore, we adopt a generative approach based on the denoising diffusion probabilistic model (DDPM)~\cite{ho2020denoising} for estimating and forecasting 3D hand and body motion.

\noindent \textbf{Conditional Diffusion Model}.
The diffusion model takes random noise sampled from Gaussian as initial motion
$x^{t}_{N} \in \mathbb{R}^{J \times 3}$
\footnote{We use superscript $^t$ for time horizon step and subscript $_n$ for diffusion denoising step.}
at temporal timestep $t$ with the maximum diffusion step $N$ and the number of joints $J$.
It iteratively removes noise at each diffusion step $n$ with a learned mean and fixed variance:
\begin{equation}
\label{eq:p_theta}
p_{\theta}(x_{n-1} \mid x_{n}, c) := \mathcal{N}(x_{n-1}; \mu_{\theta}(x_{n}, n, c), \sigma^{2}_{n}{I}),
\end{equation}
where $\mu_{\theta}(.)$ can be computed by a neural network, and $c$ is any given conditions (we detail our conditions later), to generate the 3D positions.
$\mu_{\theta}(.)$ is parameterized as follows:
\begin{equation}
\resizebox{\linewidth}{!}{
    $\mu_{\theta}(x_{n}, n, c) = \frac{\sqrt{\alpha_{n}}(1-\bar{\alpha}_{n-1})}{1-\bar{\alpha}_{n}}x_n + \frac{\sqrt{\bar{\alpha}_{n-1}}(1-\alpha_{n})}{1-\bar{\alpha}_{n}}x_{\theta}(x_n,n,c),$}
    \label{eq:mu}
\end{equation}
where $\bar{\alpha}_{n} = \prod\limits_{s=0}^{n}\alpha_{s}$, $\alpha_{n}$ is a fixed parameter.
We train the network $x_\theta(x_n, n, c)$ to directly predict $x_0$ during training time, following~\cite{li2023ego}. 
The training loss for the 3D hand and body joints prediction is defined as a reconstruction loss of the original data $x_{0}$:
\begin{equation}
\label{eq:recon-loss}
    \mathcal{L} = \mathbb{E}_{x_{0},n} \norm{{x}_{\theta}(x_n,n,c) - x_{0}}_{1}.
\end{equation}

At the inference time, 
we apply N steps of denoising with~\cref{eq:p_theta} to obtain the final denoised output $\hat x_0$.

\noindent \textbf{Conditioning Cues.}
We use three conditioning cues: i)~camera pose, ii) 2D hand locations (if visible), and iii) image features.
Humans have a remarkable ability to stabilize their heads, keeping them aligned with the body's center of mass~\cite{li2023ego}.
The camera pose, which captures the head pose given this is a head-mounted camera, provides crucial information for estimating body joints, including hand locations. 
The camera pose $c^{t}_{\text{cam}} \in \mathbb{R}^{9}$ at timestep $t$ consists of  the 6D represented rotation~\cite{zhou2019continuity} and translation vector, canonicalized w.r.t the first frame without loosing gravity information.
The left/right hand locations $(x, y)$ coordinates in 2D image space $c^{t}_{\text{left}}, c^t_{\text{right}} \in \mathbb{R}^{2}$ are utilized when they are visible to help improve forecasting accuracy.
When one hand is not visible, we replace the location of this hand side $c^t_{\text{side}}$ by $(-1, -1)$ to indicate it is invisible.
We also leverage the image features $c^{t}_{\text{img}} \in \mathbb{R}^{d_{\text{img}}}$, extracted from a visual encoder~\cite{dosovitskiy2020vit}, to provide visual context.
This complements camera poses but, importantly, allows observing any body parts in the camera view.

\noindent \textbf{Noise Encoder}.
During the observation, the diffusion model is conditioned on the above three cues, while no conditioning is used for future timesteps.
Thus, we use two types of noise encoders: conditional noise encoder $\mathcal{E}_{\text{obs}}$ shared across observation timesteps and unconditional noise encoder $\mathcal{E}_{\text{fut}}$ shared for forecasting timesteps.
The noise tokens for observation $z^{t}_{\text{obs}} \in \mathbb{R}^{d_{z}}$ at time $t$ are encoded as follows:
\begin{equation}
     z^{t}_{\text{obs}} = \mathcal{E}_{\text{obs}}(x^{t}_n, c^{t}_{\text{cam}}, c^{t}_{\text{left}}, c^t_{\text{right}}, c^{t}_{\text{img}}).
\end{equation}
As for future timesteps, these conditions cannot be obtained.
We adopt a linear layer as the noise encoder $\mathcal{E}_{\text{fut}}$ to directly tokenize the noise:
\begin{equation}
    z^{t}_{\text{fut}} = \mathcal{E}_{\text{fut}}(x^{t}_n),
\end{equation}
where $t > T$  and $z^{t}_{\text{fut}}$ 
has the same dimension as the tokens for the observation encoder.

\noindent \textbf{Denoising Transformer}.
We adopt the Transformer architecture~\cite{vaswani2017attention} as the denoising function to deal with our sequential input.
The encoded noise tokens $\bm{z}_{\text{obs}}$ and $\bm{z}_{\text{fut}}$ are concatenated and combined with the noise level information~(embedding), followed by adding positional embeddings.
The denoising function $\mathcal{T}$ takes these combined tokens as input. The final output $\hat{x}^{t}_{0}$ for each temporal timestep is obtained after passing through the decoder $\mathcal{D}_{\text{joint}}$, a linear layer, for hand and body joints:
\begin{equation}
    \hat{\bm{x}}^{1:T+F}_{0} = \mathcal{D}_{\text{joint}}(\mathcal{T}([\bm{z}^{1:T}_{\text{obs}}, \bm{z}^{T+1:T+F}_{\text{fut}}], n)).
\end{equation}

\subsection{Training Objective}
\label{sec:objective-fn}
We train our model using three losses: (1) 3D joint reconstruction loss $\mathcal{L_{\text{joint}}}$, (2) visibility loss  $\mathcal{L_{\text{vis}}}$, and (3) 2D reprojection loss for visible hands $\mathcal{L_{\text{reproj}}}$.
We detail these next.

\noindent \textbf{3D Joint Loss}.
As shown in several egocentric human body pose estimation works~\cite{li2023ego, yi2024egoallo, Luo2021DynamicsRegulatedKP, yuan2019ego}, the body pose can be estimated given the gravity-aligned camera pose.
In our work, the conditional diffusion model reconstructs 3D body joints in addition to the 3D hand poses for all timesteps: observation and forecasting.
The 3D joint loss $\mathcal{L}_{\text{joint}}$ is computed from the error between the prediction $\hat{x}_{0}$ and the ground-truth data ${x}_{0}$ using \cref{eq:recon-loss}.

\noindent \textbf{Visibility Loss}.
Furthermore, we incorporate a visibility loss so the model correctly perceives when hands are in- or out-of the field of view.
This is analogous to the visibility loss used in point tracking to address occlusion, which is one of the significant issues that lead to tracking errors.
Inspired by point tracking methods~\cite{harley2022particle, karaev23cotracker}, we incorporate the visibility loss to increase the model's ability to position hands in/out-of-view, and, importantly, regulate the learning.
We predict the visibility score $\hat v^{1:T} \in \mathbb{R}^{2T}$ for both hands in each observation timestep using the decoder $\mathcal{D}_{\text{vis}}$:
\begin{equation}
    \hat v^{1:T} = \mathcal{D}_{\text{vis}}(\mathcal{T}([\bm{z}^{1:T}_{\text{obs}}, \bm{z}^{T+1:T+F}_{\text{fut}}], n)).
\end{equation}
We employ the binary cross-entropy~(CE) loss as visibility loss $\mathcal{L}_{\text{vis}}$,
\begin{equation}
\label{eq:visibility_loss}
\mathcal{L_{\text{vis}}} = \text{CE}(v^{1:T}, \hat v^{1:T}),
\end{equation}
where the target $v^{1:T}$ can be obtained from the ground-truth.

\noindent \textbf{2D Reprojection Loss}.
We use the 2D hand location as input when the hand is visible to help improve the 3D hand pose estimation and forecasting.
However, merely adding 2D locations as input does not maintain the consistency between 2D hand input and 3D hand output.
We use the reprojection loss to penalize the error between input 2D coordinates and reprojected 2D location from the 3D output using extrinsic $\bm{P}$ and intrinsic $K$ camera parameters.
We only use the wrist position for each hand side in this reprojection loss, defined as:
\begin{equation}
    \mathcal{L_{\text{reproj}, \text{side}}} = \sum\limits_{t=1}^T v^t_{\text{side}} \norm{c^t_{\text{side}} - \Pi_{K}({\bm{P}}(\hat{x}^t_{{\text{side}}})}_{1},
\end{equation}
where $\text{side}$ is left/right, $\Pi_{K}$ denotes the projection onto 2D image space and $\hat{x}_{{\text{side}}}$ represents the reconstructed hand locations. We have $ \mathcal{L_{\text{reproj}}} =  \mathcal{L_{\text{reproj}, \text{left}}} +  \mathcal{L_{\text{reproj}, \text{right}}} $.

\noindent \textbf{Training Loss.}
We train our encoders, decoders, and denoising function with a linear combination of these losses with balancing hyperparameters for the final training loss:
\begin{equation}
    \mathcal{L} = \mathcal{L}_{\text{joint}} + \lambda_{\text{vis}}\mathcal{L}_{\text{vis}} + \lambda_{\text{reproj}}\mathcal{L}_{\text{reproj}}.
\end{equation}

\section{Experiments}
\label{sec:experiments}
In this section, we elaborate on the dataset used to train and evaluate \method (\cref{sec:dataset}), implementation details~(\cref{sec:impl}), methods we compare to (\cref{sec:comparison}), quantitative results (\cref{sec:quantitative}), ablation study of our proposed method (\cref{sec:ablation}), and qualitative results (\cref{sec:qualitative}).

\subsection{Dataset}
\label{sec:dataset}
We use the recently released Ego-Exo4D dataset, a diverse and large-scale multimodal multiview video dataset.
The dataset is released with two \textit{separate} sets of manual annotations: one for body pose (including wrist but without hand pose) and the second for hand pose.
As no prior work has targeted hand forecasting using body pose, we curate our dataset from these annotations as follows:
\begin{itemize}
    \item Ego-Exo4D Body Pose: We use the manual annotations providing 17 joints of body and wrist (without hand pose).
    \item Ego-Exo4D Hand Pose: We use the manual annotations providing $21 \times 2$ hand joints, along with automatic body annotations from exocentric cameras. We use the manual wrist annotations with the hand pose, overwriting those from the automatic body annotations. We only use the automatic body annotations for training (not evaluation).
\end{itemize}
We use the same train/val splits for both sets of annotations from~\cite{Grauman_2024_CVPR}, combining these pool of annotations to form a dataset with 3D hand and body poses even when hands are not visible from the egocentric camera.
During training with this heterogeneous data, we only backpropagate the loss on joints we have annotations for.

Aligned with prior forecasting works~\cite{ego4d, hatano2024eccvw}, we define the task so that observation is a two-second temporal duration, followed by one second of forecasting\footnote{Note that~\cite{BaoUSST_ICCV23} adopts a different protocol (0.8s obs and 0.53s forecast). We re-train this method with the standard protocol for direct comparison.}.

\cref{table: dataset} showcases the annotations we combine to form our train/test splits and those of previous datasets used to evaluate the hand forecasting.
We evaluate 15x and 9x more sequences for training and testing, respectively, compared to the H2O~\cite{Kwon_2021_H2O} dataset.
We separate sequences into those where all observation frames have in-view hands and those where one or more observation frames have out-of-view hands.
Note that these previous datasets are not only significantly smaller but also do not have 3D hand annotation when hands are out-of-view, so not suitable for our experiments.

\begin{table}[tb]
\centering
\caption{\textbf{Dataset Comparison of Train/Test Sequences}. We report the number of training and testing sequences for each dataset, H2O, EgoPAT3D, and Ego-Exo4D, categorized into in-view, out-of-view scenarios, and total sequences. The sequence counts are provided separately for each hand side. Moreover, we report the availability of body pose annotation.}
\vspace*{-8pt}
\centering
\scalebox{0.6}{
\begin{tabular}{lccccccc}
\toprule
\multirow{2}{*}{Dataset} & \multirow{2}{*}{Body Pose} & \multicolumn{2}{c}{In-view} & \multicolumn{2}{c}{Out-of-view} & \multicolumn{2}{c}{All}\\
\cmidrule(l{2pt}r{3pt}){3-4} \cmidrule(l{2pt}r{3pt}){5-6} \cmidrule(l{2pt}r{3pt}){7-8}
 & &
 \multicolumn{1}{r}{train} & \multicolumn{1}{l}{test} & 
 \multicolumn{1}{r}{train} & \multicolumn{1}{l}{test} &
 \multicolumn{1}{r}{train} & \multicolumn{1}{l}{test} \\
\toprule
H2O~\cite{Kwon_2021_H2O} & & 9.9k & 3.7k & - & - & 9.9k & 3.7k \\
EgoPAT3D~\cite{Li_2022_EGOPAT} &  & 7.2k & 3.8k & - & - & 7.2k & 3.8k \\
\rowcolor{lightgray}
Ego-Exo4D~\cite{Grauman_2024_CVPR} (Body) & \checkmark & 52.4k & 11.6k & 85.2k & 18.9k &  138k & 30.5k\\
\rowcolor{lightgray}
Ego-Exo4D~\cite{Grauman_2024_CVPR} (Hand) & \checkmark & 14.2k & 3.4k & 4.5k & 0.1k &  18.7k & 3.5k\\

Ego-Exo4D~\cite{Grauman_2024_CVPR} (Ours) & \checkmark & 66.6k & 15.0k & 89.7k & 19.0k &  156k & 34.0k\\
\bottomrule
\end{tabular}}
\vspace*{-8pt}
\label{table: dataset}
\end{table}

\subsection{Implementation Details}
\label{sec:impl}

\begin{table*}[tb]
\centering
\caption{\textbf{Hand Forecasting Accuracy}. We report the hand trajectory and pose forecasting results on in-view, out-of-view, and all scenarios on the Ego-Exo4D dataset.}
\vspace*{-12pt}
\centering
\scalebox{0.8}{
\begin{tabular}{l cccccc c cccccc}
\toprule
\multirow{3}{*}{Method} & \multicolumn{6}{c}{Hand Trajectory Forecasting} & & \multicolumn{6}{c}{Hand Pose Forecasting}\\
\cmidrule(l{2pt}r{3pt}){2-7} \cmidrule(l{2pt}r{3pt}){9-14}
& \multicolumn{2}{c}{In-view} & \multicolumn{2}{c}{Out-of-view} & \multicolumn{2}{c}{All} & & \multicolumn{2}{c}{In-view} & \multicolumn{2}{c}{Out-of-view} & \multicolumn{2}{c}{All}\\
\cmidrule(l{2pt}r{3pt}){2-3} \cmidrule(l{2pt}r{3pt}){4-5} \cmidrule(l{2pt}r{3pt}){6-7} \cmidrule(l{2pt}r{3pt}){9-10} \cmidrule(l{2pt}r{3pt}){11-12} \cmidrule(l{2pt}r{3pt}){13-14}
 & ADE & FDE & ADE & FDE & ADE & FDE & & MPJPE & MPJPE-F & MPJPE & MPJPE-F & MPJPE & MPJPE-F\\
\toprule
Static & 0.199 & 0.209 & 0.434 & 0.546 & 0.335 & 0.405 & & 0.163 & 0.176 & 0.297 & 0.325 & 0.166 & 0.179\\
CVM~\cite{cvm} & 0.201 & 0.217 & 0.451 & 0.648 & 0.346 & 0.467 & & 0.162 & 0.177 & 0.352 & 0.427 & 0.166 & 0.183 \\
EgoEgoForecast & 0.171 & 0.185 & 0.385 & 0.472 & 0.295 & 0.352 & & 0.162 & 0.173 & 0.299 & 0.345 & 0.166 & 0.177\\
USST~\cite{BaoUSST_ICCV23} & 0.277 & 0.280 & 0.763 & 0.792  & 0.562 & 0.581 & & - & - & - & - & - & - \\
\cmidrule(l){1-14} 
\rowcolor{lightgray}
Ours & \textbf{0.116} & \textbf{0.152} & \textbf{0.366} & \textbf{0.459} & \textbf{0.261} & \textbf{0.324} & & \textbf{0.112} & \textbf{0.140} & \textbf{0.240} & \textbf{0.280} & \textbf{0.115} & \textbf{0.143}\\
\bottomrule
\end{tabular}}
\vspace*{-12pt}
\label{table: main-forecast}
\end{table*}

\noindent \textbf{Experimental Setup}.
We sample at 10 FPS (frames per second) for observation and forecasting.
As a result, we have $T=20$ sampled frames for input observation and $F=10$ sampled frames for forecasting.
Visual features are extracted from a pre-trained on ImageNet and frozen ViT-S.
We reconstruct the hand and body joints, with the number of joints $J = 57$ -- ($15$ body joints exc. wrist + $21 \times 2$ hand joints).
We use the ground-truth 2D hand locations for input, normalized to the range of~$[0,1]$.

\noindent \textbf{Training}.
We train the model from random weights for 40K iterations with a constant learning rate of $1e-4$.
Regarding the parameters for the objective function, we empirically choose each balancing hyperparameter $\lambda_{\text{vis}}$ and $\lambda_{\text{reproj}}$ to $1e-1$ and $5e-2$, respectively.
(See suppl. for ablation).

\noindent \textbf{Evaluation Metrics}.
We report the Average Displacement Error~(ADE) and Final Displacement Error~(FDE) in global 3D space for wrist trajectory forecasting, often used in trajectory forecasting works. 
Regarding hand pose forecasting, we adopt the Mean Per Joint Position Error~(MPJPE), which averages all future timesteps, and MPJPE-F, which averages the performance at the last forecasting frame.
Furthermore, we report MPJPE and Mean Per Joint Velocity Error~(MPJVE) to evaluate the accuracy of body pose estimation and forecasting.
All 3D evaluation metrics are reported in meters.
We generate one sample during evaluation for fair comparisons with the existing deterministic models.

\subsection{Baselines}
\label{sec:comparison}
We use two naive baselines and three previous works (one for body pose estimation and one adapted to forecasting):
\begin{itemize}
    \item \textbf{Static} is a naive baseline that keeps the average whole-body pose of training data at the last observable timestep. It showcases the difficulty of the dataset.
    \item \textbf{CVM~\cite{cvm}} is another naive baseline that is often used in trajectory forecasting. The Constant Velocity Model~(CVM) assumes that the most recent relative velocity is the most relevant predictor for future trajectory.
    \item \textbf{EgoAllo~\cite{yi2024egoallo}} is a diffusion-based model for body pose estimation. The method is not designed for forecasting and does not attempt it. We evaluate the guidance-free version of EgoAllo,  pre-trained on AMASS~\cite{mahmood2019amass} dataset\footnote{We use the SMPL's internal joint regressor to convert into MS COCO 17 body joints.}.
    \item \textbf{EgoEgoForecast (Baseline)} We extend~\cite{li2023ego} to a forecasting method. Similar to our proposed method, there are two noise encoders: one is for encoding with head poses during observation, and the other is for encoding random noise.
    This model can be seen as an architectural baseline for our \method, as it is a vanilla diffusion model without the additional conditioning losses we introduce. The model is trained from scratch on Ego-Exo4D dataset.
    \item \textbf{USST~\cite{BaoUSST_ICCV23}} is the only prior work that evaluates egocentric 3D hand trajectory forecasting. 
    We retrained USST on our dataset using the official implementation.
    When the hand is out-of-view during observation, we use masking for both training and inference.
\end{itemize}

\begin{table}[tb]
\centering
\caption{\textbf{Architecture and Losses Ablations}. $\mathcal{L}_{\text{body}}$ represents the reconstruction loss for body joints. The $\mathcal{L}_{\text{obs}}$ represents all losses in the observation timesteps, including 3D joint loss during observation, reprojection loss, and visibility loss.}
\vspace*{-8pt}
\centering
\scalebox{0.64}{
\begin{tabular}{lcccccc}
\toprule
\multirow{2}{*}{Method} & \multicolumn{3}{c}{Hand Trajectory Forecasting} & \multicolumn{3}{c}{Hand Pose Forecasting}\\
\cmidrule(l{2pt}r{3pt}){2-4} \cmidrule(l{2pt}r{3pt}){5-7}
 & In-view & Out-of-view & All & In-view & Out-of-view & All\\
\toprule
EgoEgoForecast & 0.171 & 0.385 & 0.295 & 0.162 & 0.299 & 0.166 \\
\cmidrule(l){1-7} 
Ours w/o. 2D joint & 0.151 & 0.377 & 0.282 & 0.139 & 0.269 & 0.142\\
Ours w/o. image & \textbf{0.116} & 0.367 & \textbf{0.261} & 0.117 & \textbf{0.234} & 0.120 \\
Ours w/o. $\mathcal{L}_{\text{reproj}}$ & 0.132 & 0.368 & 0.269 & 0.125 & 0.250 & 0.128 \\
Ours w/o. $\mathcal{L}_{\text{vis}}$ & 0.127 & 0.377 & 0.272 & 0.121 & 0.240 & 0.124 \\
Ours w/o. $\mathcal{L}_{\text{body}}$ & 0.129 & 0.385 & 0.277 & 0.120 & 0.258 & 0.123 \\
Ours w/o. $\mathcal{L}_{\text{obs}}$ & 0.149 & 0.390 & 0.289 & 0.139 & 0.250 & 0.142\\
\cmidrule(l){1-7} 
\rowcolor{lightgray}
Ours & \textbf{0.116} & \textbf{0.366} & \textbf{0.261} & \textbf{0.112} & 0.240 & \textbf{0.115} \\
\bottomrule
\end{tabular}}
\label{table: input-loss-ablation}
\end{table}

\begin{table}[tb]
\centering
\caption{\textbf{Evaluation on Different Out-of-view Ratio Intervals}. We report the ADE and FDE results across five equally divided out-of-view ratio intervals $\gamma_{(i, j]}$ ranging from zero to one.}
\centering
\vspace*{-10pt}
\resizebox{\linewidth}{!}{
\begin{tabular}{l cc cc cc cc cc}
\toprule 
\multirow{2}{*}{Method} &
\multicolumn{2}{c}{$\gamma_{(0.0,0.2]}$} &
\multicolumn{2}{c}{$\gamma_{(0.2,0.4]}$} &
\multicolumn{2}{c}{$\gamma_{(0.4,0.6]}$} &
\multicolumn{2}{c}{$\gamma_{(0.6,0.8]}$} &
\multicolumn{2}{c}{$\gamma_{(0.8,1.0]}$} \\
\cmidrule(l{2pt}r{3pt}){2-3}
\cmidrule(l{2pt}r{3pt}){4-5}
\cmidrule(l{2pt}r{3pt}){6-7}
\cmidrule(l{2pt}r{3pt}){8-9}
\cmidrule(l{2pt}r{3pt}){10-11}
  & 
\scalebox{1}{ADF} & \scalebox{1}{FDE} &
\scalebox{1}{ADF} & \scalebox{1}{FDE} &
\scalebox{1}{ADF} & \scalebox{1}{FDE} &
\scalebox{1}{ADF} & \scalebox{1}{FDE} &
\scalebox{1}{ADF} & \scalebox{1}{FDE} \\
\toprule
EgoEgoForecast &
0.284 & 0.329 &
0.393 & 0.475 &
0.424 & 0.519 &
0.415 & 0.504 &
0.363 & 0.459 \\
Ours w/o. $\mathcal{L}_{\text{body}}$ &
0.250 & 0.300 &
0.398 & 0.477 &
0.432 & 0.519 &
0.430 & 0.514 &
0.353 & 0.442 \\
Ours w/o. $\mathcal{L}_{\text{vis}}$ &
0.254 & 0.303 &
0.394 & 0.487 &
0.423 & 0.521 &
0.412 & \textbf{0.498} &
0.349 & 0.447\\
\cmidrule(l){1-11} 
\rowcolor{lightgray}
Ours &
\textbf{0.236} & \textbf{0.284} &
\textbf{0.379} & \textbf{0.476} &
\textbf{0.417} & \textbf{0.517} &
\textbf{0.404} & 0.505 &
\textbf{0.335} & \textbf{0.434} \\
\bottomrule
\end{tabular}}
\vspace*{-12pt}
\label{table: oov-ablation}
\end{table}

\begin{table}[tb]
\centering
\caption{\textbf{Body Pose Estimation/Forecasting Accuracy}. Comparison with the body pose estimation/forecasting in terms of MPJPE and MPJVE. The location-based model, used as a baseline of the body pose estimation in the Ego-Exo4D, is a transformer that takes head poses as input to output the body joints. * denotes the method is not trained on the Ego-Exo4D.}
\centering
\scalebox{0.6}{
\begin{tabular}{lcccc}
\toprule
\multirow{2}{*}{Method} &
\multicolumn{2}{c}{Observation} & \multicolumn{2}{c}{Forecasting}\\
\cmidrule(l{2pt}r{3pt}){2-3} \cmidrule(l{2pt}r{3pt}){4-5} 
  & MPJPE & MPJVE & MPJPE & MPJVE \\
\toprule
Static & 0.357 & 0.778 & 0.286 & 0.778 \\
Location-based~\cite{Grauman_2024_CVPR} & 0.148 & \textbf{0.583} & - & - \\
EgoAllo~\cite{yi2024egoallo}$^*$ & 0.219 & - & - & - \\
EgoEgoForecast & 0.173 & 0.651 & 0.245 & 0.771 \\
\cmidrule(l){1-5} 
\rowcolor{lightgray}
Ours & \textbf{0.142} & 0.697 & \textbf{0.221} & \textbf{0.763} \\
\bottomrule
\end{tabular}}
\label{table: body-pose-accuracy}
\end{table}

\begin{table}[tb]
\centering
\caption{\textbf{Hand Pose Estimation Accuracy}. Comparison with the hand pose estimation in terms of MPJPE.}
\centering
\scalebox{0.7}{
\begin{tabular}{lccc}
\toprule
Method & In-view & Out-of-view & All\\
\toprule
THOR-net~\cite{Aboukhadra_2023_WACV} & 0.051 & - & - \\
POTTER~\cite{zheng2023potter} & \textbf{0.031} & - & - \\
EgoEgoForecast & 0.158 & 0.289 & 0.161 \\
\cmidrule(l){1-4} 
\rowcolor{lightgray}
Ours & 0.067 & \textbf{0.206} & \textbf{0.081} \\
\bottomrule
\end{tabular}}
\vspace*{-12pt}
\label{table: hand-pose-accuracy}
\end{table}

\subsection{Quantitative Results}
\label{sec:quantitative}
\noindent \textbf{Hand Trajectory and Pose Forecasting}.
We compare the performance of egocentric 3D hand trajectory forecasting with the noted baselines above on the Ego-Exo4D dataset.
We report ADE and FDE in two different scenarios: 1) \textit{in-view} scenario where a hand is in-view in all observation timesteps, same as the previous evaluation setup, and 2) \textit{out-of-view} scenario where a hand is out of the field of view at least one timestep during observation.
We consider the out-of-view sample for each hand side: the same sequence can be in-view for left hand but out-of-view for right hand.
We show the results in \cref{table: main-forecast}.

Naive baselines that assume no motion (i.e., static position) or constant velocity fail to capture the complex dynamics of future hand movements.
This shows the challenging aspect of the 3D hand forecasting task and this dataset.
USST, poorly performs especially when hands are out-of-view. It indicates that 2D hand + 3D camera pose is more suitable as input than 3D hand alone, as the latter is unavailable when hands are out-of-view.

Our proposed method, \method, outperforms the baseline EgoEgoForecast, which forecasts hands and body joints, across both in-view and out-of-view scenarios.
Notably, significant improvements are observed in the in-view scenario, as the proposed model incorporates visual cues during observation, such as image features and 2D hand locations, which the baseline lacks.
Additionally, while EgoEgoForecast leverages body movement information to enable forecasting for out-of-view scenarios, our model further improves on this by introducing awareness of in-view and out-of-view status for each hand side during observation, resulting in superior forecasting accuracy.

We also evaluate the hand pose forecasting performance except for USST as the method cannot predict multiple joints. \method outperforms baselines on in- and out-of-view scenarios. The improvements over EgoEgoForecast support the utilization of visible cues for pose forecasting.

\subsection{Ablation Analysis}
\label{sec:ablation}
\noindent \textbf{Architecture and Losses Ablation}.
This ablation study focuses on the conditional input and loss components to verify the contribution of each module.
We evaluate the hand trajectory and pose forecasting in \cref{table: input-loss-ablation}.
We experiment with removing each component individually and compare with the baseline EgoEgoForecast and the full model of \method, including 1) visible 2D hand joints, 2) image features, 3) 2D reprojection loss $\mathcal{L}_{\text{reproj}}$, 4) visibility loss $\mathcal{L}_{\text{vis}}$, 5) body pose loss $\mathcal{L}_{\text{body}}$, and 6) losses during observation $\mathcal{L}_{\text{obs}}$.

Without 2D hand joint coordinates, significant performance drops can be seen in both in-view and out-of-view sequences.
In contrast, the image features only marginally improve performance and marginally harm performance for out-of-view hand pose forecasting.
The 2D reprojection loss serves as an effective regularization, further boosting hand forecasting performance and underscoring the importance of maintaining spatial consistency between observed and predicted hand positions in 2D image space. 
Notably, without our proposed visibility loss or body pose loss, hand trajectory forecasting performance degrades significantly in out-of-view scenarios.
Lastly, estimating and optimizing joints for the observation timestamps $\mathcal{L}_{obs}$, instead of solely optimizing 3D joints during the forecasting period, also improves results for trajectory and pose forecasting.

Overall, our full model performs the best among the variants of our model in the entire evaluation set.

\begin{figure*}[t]
\centering
   \includegraphics[width=0.85\linewidth]{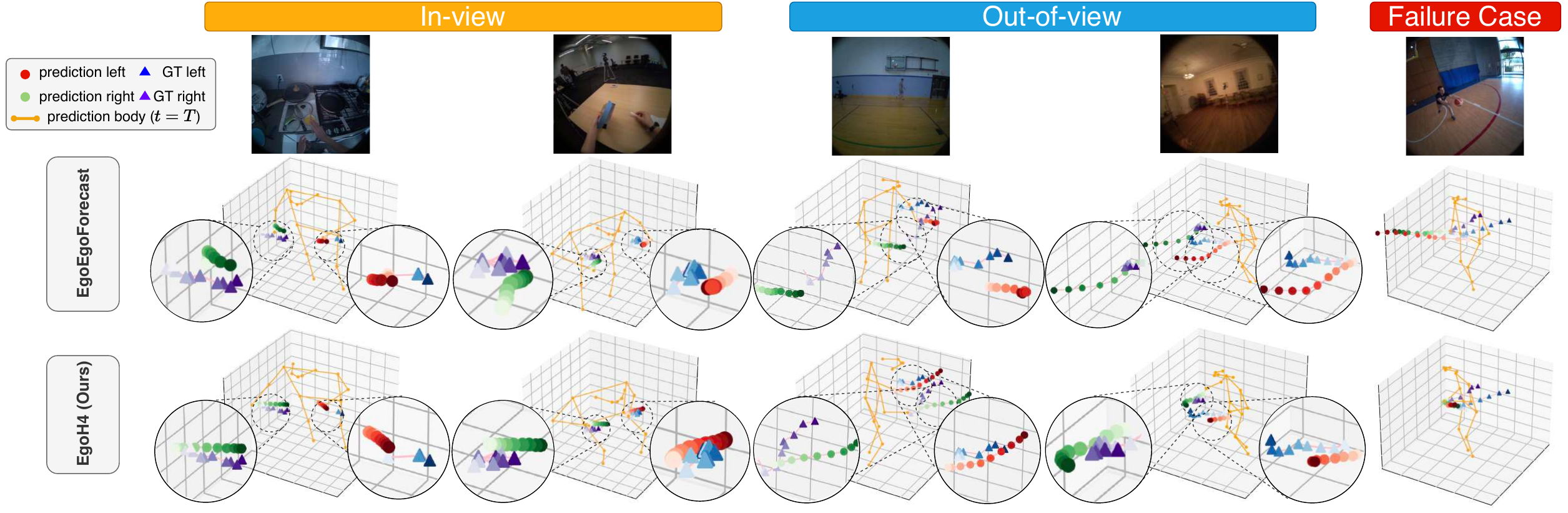}
   \caption{\textbf{Qualitative results for hand trajectory forecasting.} We show sample qualitative results compared to our best-performing baseline across activities: cooking, covid testing, basketball, and dance exercises. Dots in \textcolor{red}{red}, \textcolor{ForestGreen}{green}, \textcolor{blue}{blue}, \textcolor{Purple}{purple}, and \textcolor{BurntOrange}{orange} represent the prediction of left/right future hands, ground-truth of left/right hands, and the prediction of body joints at the last observable frame, respectively. For each track, darker colors indicate later times.} 
\label{fig:qualitative}
   \includegraphics[width=0.8\linewidth]{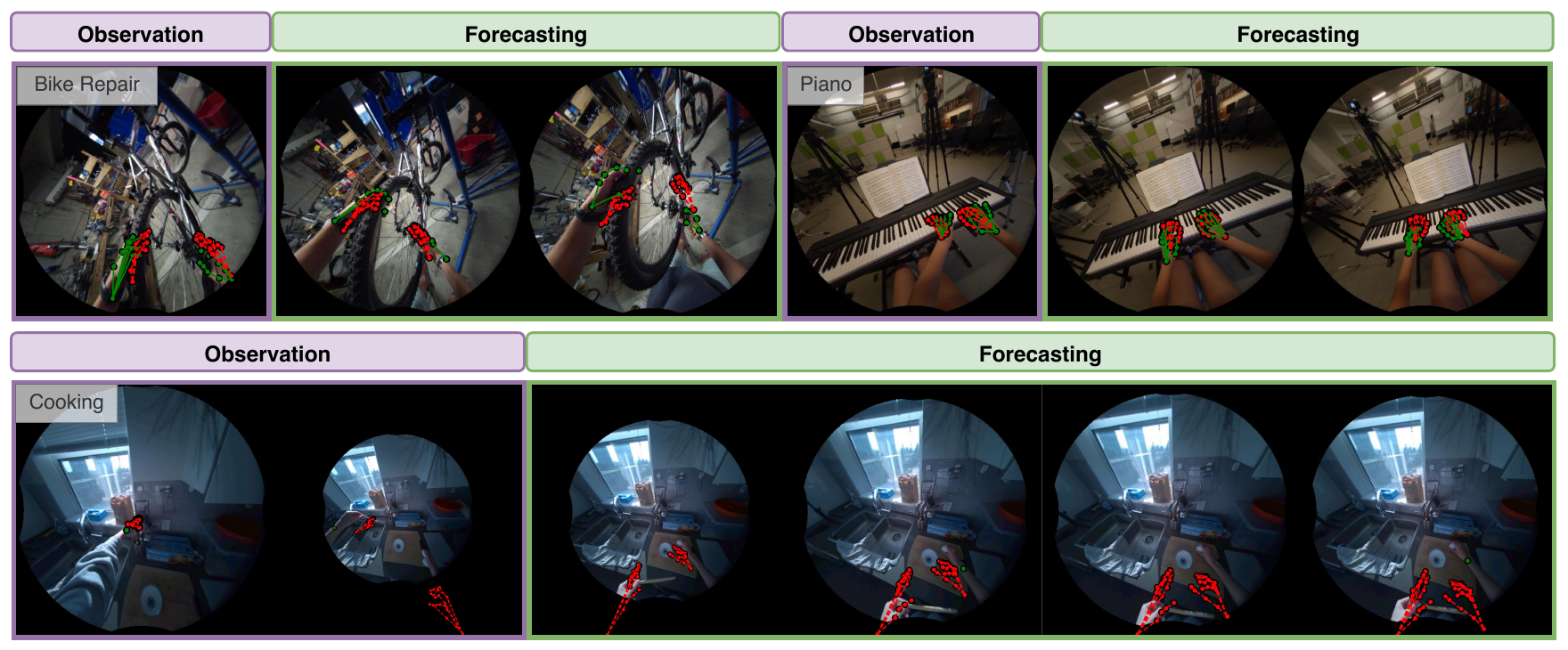}
   \caption{\textbf{Qualitative results for hand pose forecasting.} We show two in-view cases, a bike repair and piano playing scene, in the first row and one out-of-view case, a cooking scene, in the second row. Dots in \textcolor{red}{red} and \textcolor{ForestGreen}{green} denote the prediction and ground truth, respectively. Note that we expand the image plane so that we can also show the out-of-view hands.} 
   \vspace*{-12pt}
\label{fig:qualitative-pose}
\end{figure*}

\noindent \textbf{Impact of Out-of-view Ratio During Observation}.
We conduct an in-depth analysis to better understand performance as the ratio of hands out-of-view varies during observation.
Let $\gamma$ represent the out-of-view ratio, calculated as $\frac{h}{T}$, where $h$ is the number of out-of-view frames, and $T$ is the total number of observation frames.
To systematically analyze the performance of models across varying out-of-view ratios, we divide $\gamma$ into discrete intervals.
We define the interval $\gamma_{(i,j]}$ as follows: $\gamma_{(i,j]}: 0 \le i < \gamma \le j \le 1$.
We compare the forecasting accuracy with the baseline and variants of our models: without the visibility loss $\mathcal{L}_{\text{vis}}$ and the body pose loss $\mathcal{L}_{\text{body}}$ since these loss components are effective for out-of-view cases.

As shown in \cref{table: oov-ablation}, the proposed method consistently outperforms the baseline and the variants of our models across all out-of-view ratio intervals in both ADE and FDE.
Without jointly optimizing with body poses or visibility loss, hand trajectory forecasting performance drops by an average of 1.8/1.2 cm regardless of the out-of-view ratio, respectively.
These results suggest that both jointly optimizing body pose and visibility awareness are crucial for enhancing hand forecasting accuracy.

\noindent \textbf{Body Pose Estimation/Forecasting}.
To assist 3D hand forecasting, our method predicts body pose jointly.
Here we compare our body pose estimation with other relevant body estimation methods during both observation and forecasting.
As shown in \cref{table: body-pose-accuracy}, our proposed model surpasses the baseline.
MPJPE is improved by 3cm over the EgoEgoForecast when estimating body pose from given head poses, and by 2.4cm in forecasting future body pose.
The performance of EgoAllo~\cite{yi2024egoallo}, pretrained on AMASS, suggests that the training on body motion data from the in-the-wild dataset is necessary.

\noindent \textbf{Hand Pose Estimation}.
We report the hand pose estimation performance during observation\footnote{We report the wrist-relative MPJPE for in-view sequences, following Ego-Exo4D evaluation.} in ~\cref{table: hand-pose-accuracy}. We compare with EgoEgoForecast as well as the baseline used as the hand pose estimation task in the Ego-Exo4D paper: THOR-net~\cite{Aboukhadra_2023_WACV} and POTTER~\cite{zheng2023potter}. 
Note that these methods only optimize for observed frames, and thus a drop in performance of \method when attempting forecasting would be expected.
\method outperforms EgoEgoForecast by a large margin.

\subsection{Qualitative Results}
\label{sec:qualitative}

The qualitative results for hand trajectory forecasting on the Ego-Exo4D dataset in different scenarios, in- or out-of-view, are visualized in \cref{fig:qualitative}.
The in-view results demonstrate that \method accurately forecasts the hand locations compared to the baseline as we leverage visible cues as conditions.
In out-of-view scenarios, our model effectively predicts future hand trajectories. 
In \cref{fig:qualitative}, the failure case shows the ground truth body pose moving backward, while the body poses are forecasted to stay around. This reveals a limitation: significant errors in body pose forecasting can negatively impact hand forecasting accuracy.

We visualize hand pose forecasting results, reprojected on the 2D image, in \cref{fig:qualitative-pose}. The first row demonstrates our method's ability to estimate and forecast the hand joints correctly. The second row shows an interesting case where the right hand comes into view during observation. The right hand is forecasted to take a bowl and manipulate it with two hands. This shows the capability of our approach to deal with out-of-view scearios by leveraging the body motion.

\vspace*{-8pt}
\section{Conclusion}
\label{sec:conclusion}
\vspace*{-4pt}

\noindent \textbf{Conclusion}.
We are the first to explore egocentric 3D hand position and pose forecasting even when hands are partially or completely invisible during observation.
We propose \method, a diffusion-based transformer model that denoises the body and hand joints, given head pose, 2D hand locations (if visible), and image features.
Leveraging knowledge of body motion enhances our method of estimating/forecasting the invisible hand and improves hand forecasting accuracy.
Moreover, we employ the 3D-to-2D reprojection loss for prediction consistency and visibility loss to acquire out-of-view awareness.
We evaluate our proposed method on the Ego-Exo4D dataset, showing significantly improved forecasting accuracy for the in- and out-of-view sequences.

\noindent \textbf{Limitations and Future Work}.
When the hands are invisible, our model primarily relies on body joint information to estimate and forecast the 3D hand position. 
This can be ambiguous -- with the same input, the hands can be in different positions - e.g., by the side of the body or behind one's back. 
While the diffusion model is capable of generating both options, performance degrades when the distribution of body movements during evaluation differs from those of training.
This is a known limitation in evaluating forecasting, and we leave its exploration to future work.

\section*{Acknowledgements}
M Hatano is supported by JSPS Overseas Challenge Program for Young Researchers, JST BOOST, Japan Grant Number JPMJBS2409, and Amano Institute of Technology.
Z Zhu is supported by University of Bristol - Chinese Scholarship Council studentship.
D Damen is supported by EPSRC Fellowship UMPIRE (EP/T004991/1).
{
    \small
    \bibliographystyle{ieeenat_fullname}
    \bibliography{main}

\begin{thebibliography}{83}
\providecommand{\natexlab}[1]{#1}
\providecommand{\url}[1]{\texttt{#1}}
\expandafter\ifx\csname urlstyle\endcsname\relax
  \providecommand{\doi}[1]{doi: #1}\else
  \providecommand{\doi}{doi: \begingroup \urlstyle{rm}\Url}\fi

\bibitem[Aboukhadra et~al.(2023)Aboukhadra, Malik, Elhayek, Robertini, and Stricker]{Aboukhadra_2023_WACV}
Ahmed~Tawfik Aboukhadra, Jameel Malik, Ahmed Elhayek, Nadia Robertini, and Didier Stricker.
\newblock Thor-net: End-to-end graformer-based realistic two hands and object reconstruction with self-supervision.
\newblock In \emph{WACV}, 2023.

\bibitem[Akada et~al.(2022)Akada, Wang, Shimada, Takahashi, Theobalt, and Golyanik]{hakada2022unrealego}
Hiroyasu Akada, Jian Wang, Soshi Shimada, Masaki Takahashi, Christian Theobalt, and Vladislav Golyanik.
\newblock Unrealego: A new dataset for robust egocentric 3d human motion capture.
\newblock In \emph{ECCV}, 2022.

\bibitem[Bao et~al.(2023)Bao, Chen, Zeng, Li, Xu, Yuan, and Kong]{BaoUSST_ICCV23}
Wentao Bao, Lele Chen, Libing Zeng, Zhong Li, Yi Xu, Junsong Yuan, and Yu Kong.
\newblock Uncertainty-aware state space transformer for egocentric 3d hand trajectory forecasting.
\newblock In \emph{ICCV}, 2023.

\bibitem[Cao et~al.(2020)Cao, Gao, Mangalam, Cai, Vo, and Malik]{cao2020long}
Zhe Cao, Hang Gao, Karttikeya Mangalam, Qi-Zhi Cai, Minh Vo, and Jitendra Malik.
\newblock Long-term human motion prediction with scene context.
\newblock In \emph{ECCV}, 2020.

\bibitem[Castillo et~al.(2023)Castillo, Escobar, Jeanneret, Pumarola, Arbel{\'a}ez, Thabet, and Sanakoyeu]{castillo2023bodiffusion}
Angela Castillo, Maria Escobar, Guillaume Jeanneret, Albert Pumarola, Pablo Arbel{\'a}ez, Ali Thabet, and Artsiom Sanakoyeu.
\newblock Bodiffusion: Diffusing sparse observations for full-body human motion synthesis.
\newblock In \emph{ICCV}, 2023.

\bibitem[Chen et~al.(2023)Chen, Zhang, Li, Pang, Xia, and Liu]{chen2023humanmac}
Ling-Hao Chen, Jiawei Zhang, Yewen Li, Yiren Pang, Xiaobo Xia, and Tongliang Liu.
\newblock Humanmac: Masked motion completion for human motion prediction.
\newblock In \emph{ICCV}, 2023.

\bibitem[Cheng et~al.(2023)Cheng, Shan, Hassen, Higgins, and Fouhey]{cheng23_handv2}
Tianyi Cheng, Dandan Shan, Ayda Hassen, Richard Higgins, and David Fouhey.
\newblock Towards a richer 2d understanding of hands at scale.
\newblock In \emph{NeurIPS}, 2023.

\bibitem[Chi et~al.(2024)Chi, Huang, Sachdeva, Ma, Ramani, and Lee]{chi2025estimating}
Seunggeun Chi, Pin-Hao Huang, Enna Sachdeva, Hengbo Ma, Karthik Ramani, and Kwonjoon Lee.
\newblock Estimating ego-body pose from doubly sparse egocentric video data.
\newblock In \emph{NeurIPS}, 2024.

\bibitem[Choudhury et~al.(2023)Choudhury, Kitani, and Jeni]{choudhury2023tempo}
Rohan Choudhury, Kris~M Kitani, and L{\'a}szl{\'o}~A Jeni.
\newblock Tempo: Efficient multi-view pose estimation, tracking, and forecasting.
\newblock In \emph{ICCV}, 2023.

\bibitem[Contributors(2020)]{mmpose2020}
MMPose Contributors.
\newblock Openmmlab pose estimation toolbox and benchmark.
\newblock \url{https://github.com/open-mmlab/mmpose}, 2020.

\bibitem[Cuevas-Velasquez et~al.(2024)Cuevas-Velasquez, Hewitt, Aliakbarian, and Baltru{\v{s}}aitis]{cuevas2024simpleego}
Hanz Cuevas-Velasquez, Charlie Hewitt, Sadegh Aliakbarian, and Tadas Baltru{\v{s}}aitis.
\newblock Simpleego: Predicting probabilistic body pose from egocentric cameras.
\newblock In \emph{3DV}, 2024.

\bibitem[Darkhalil et~al.(2022)Darkhalil, Shan, Zhu, Ma, Kar, Higgins, Fidler, Fouhey, and Damen]{VISOR2022}
Ahmad Darkhalil, Dandan Shan, Bin Zhu, Jian Ma, Amlan Kar, Richard Higgins, Sanja Fidler, David Fouhey, and Dima Damen.
\newblock Epic-kitchens visor benchmark: Video segmentations and object relations.
\newblock In \emph{NeurIPS}, 2022.

\bibitem[Diller et~al.(2024)Diller, Funkhouser, and Dai]{diller2023futurehuman3d}
Christian Diller, Thomas Funkhouser, and Angela Dai.
\newblock Futurehuman3d: Forecasting complex long-term 3d human behavior from video observations.
\newblock In \emph{CVPR}, 2024.

\bibitem[Doosti et~al.(2020)Doosti, Naha, Mirbagheri, and Crandall]{doosti_2020_cvpr}
Bardia Doosti, Shujon Naha, Majid Mirbagheri, and David Crandall.
\newblock Hope-net: A graph-based model for hand-object pose estimation.
\newblock In \emph{CVPR}, 2020.

\bibitem[Dosovitskiy et~al.(2021)Dosovitskiy, Beyer, Kolesnikov, Weissenborn, Zhai, Unterthiner, Dehghani, Minderer, Heigold, Gelly, Uszkoreit, and Houlsby]{dosovitskiy2020vit}
Alexey Dosovitskiy, Lucas Beyer, Alexander Kolesnikov, Dirk Weissenborn, Xiaohua Zhai, Thomas Unterthiner, Mostafa Dehghani, Matthias Minderer, Georg Heigold, Sylvain Gelly, Jakob Uszkoreit, and Neil Houlsby.
\newblock An image is worth 16x16 words: Transformers for image recognition at scale.
\newblock \emph{ICLR}, 2021.

\bibitem[Engel et~al.(2023)Engel, Somasundaram, Goesele, Sun, Gamino, Turner, Talattof, Yuan, Souti, Meredith, Peng, Sweeney, Wilson, and et~al.]{engel2023projectarianewtool}
Jakob Engel, Kiran Somasundaram, Michael Goesele, Albert Sun, Alexander Gamino, Andrew Turner, Arjang Talattof, Arnie Yuan, Bilal Souti, Brighid Meredith, Cheng Peng, Chris Sweeney, Cole Wilson, and et al.
\newblock Project aria: A new tool for egocentric multi-modal ai research.
\newblock \emph{arXiv preprint arXiv:2308.13561}, 2023.

\bibitem[Escobar et~al.(2025)Escobar, Puentes, Forigua, Pont-Tuset, Maninis, and Arbeláez]{escobar2025egocast}
Maria Escobar, Juanita Puentes, Cristhian Forigua, Jordi Pont-Tuset, Kevis-Kokitsi Maninis, and Pablo Arbeláez.
\newblock Egocast: Forecasting egocentric human pose in the wild.
\newblock In \emph{WACV}, 2025.

\bibitem[Fan et~al.(2024)Fan, Ohkawa, Yang, Lin, Zhou, Zhou, Liang, Gao, Zhang, Zhang, et~al.]{fan2025benchmarks}
Zicong Fan, Takehiko Ohkawa, Linlin Yang, Nie Lin, Zhishan Zhou, Shihao Zhou, Jiajun Liang, Zhong Gao, Xuanyang Zhang, Xue Zhang, et~al.
\newblock Benchmarks and challenges in pose estimation for egocentric hand interactions with objects.
\newblock In \emph{ECCV}, 2024.

\bibitem[Furnari and Farinella(2019)]{furnari2019would}
Antonino Furnari and Giovanni~Maria Farinella.
\newblock What would you expect? anticipating egocentric actions with rolling-unrolling lstms and modality attention.
\newblock In \emph{ICCV}, 2019.

\bibitem[Garcia-Hernando et~al.(2018)Garcia-Hernando, Yuan, Baek, and Kim]{FirstPersonAction_CVPR2018}
Guillermo Garcia-Hernando, Shanxin Yuan, Seungryul Baek, and Tae-Kyun Kim.
\newblock First-person hand action benchmark with rgb-d videos and 3d hand pose annotations.
\newblock In \emph{CVPR}, 2018.

\bibitem[Girdhar and Grauman(2021)]{girdhar2021anticipative}
Rohit Girdhar and Kristen Grauman.
\newblock Anticipative video transformer.
\newblock In \emph{ICCV}, 2021.

\bibitem[Goel et~al.(2023)Goel, Pavlakos, Rajasegaran, Kanazawa, and Malik]{goel2023humans}
Shubham Goel, Georgios Pavlakos, Jathushan Rajasegaran, Angjoo Kanazawa, and Jitendra Malik.
\newblock Humans in 4{D}: Reconstructing and tracking humans with transformers.
\newblock In \emph{ICCV}, 2023.

\bibitem[Goletto et~al.(2024)Goletto, Nagarajan, Averta, and Damen]{goletto2024amego}
Gabriele Goletto, Tushar Nagarajan, Giuseppe Averta, and Dima Damen.
\newblock Amego: Active memory from long egocentric videos.
\newblock In \emph{ECCV}, 2024.

\bibitem[Goyal et~al.(2022)Goyal, Modi, Goyal, and Gupta]{goyal2022human}
Mohit Goyal, Sahil Modi, Rishabh Goyal, and Saurabh Gupta.
\newblock Human hands as probes for interactive object understanding.
\newblock In \emph{CVPR}, 2022.

\bibitem[Grauman et~al.(2022)Grauman, Westbury, Byrne, Chavis, Furnari, Girdhar, Hamburger, Jiang, Liu, Liu, Martin, Nagarajan, Radosavovic, Ramakrishnan, Ryan, Sharma, Wray, Xu, Xu, Zhao, Bansal, Batra, Cartillier, Crane, Do, Doulaty, Erapalli, Feichtenhofer, Fragomeni, Fu, Gebreselasie, Gonz\'alez, Hillis, Huang, Huang, Jia, Khoo, Kol\'a\v{r}, Kottur, Kumar, Landini, Li, Li, Li, Mangalam, Modhugu, Munro, Murrell, Nishiyasu, Price, Ruiz, Ramazanova, Sari, Somasundaram, Southerland, Sugano, Tao, Vo, Wang, Wu, Yagi, Zhao, Zhu, Arbel\'aez, Crandall, Damen, Farinella, Fuegen, Ghanem, Ithapu, Jawahar, Joo, Kitani, Li, Newcombe, Oliva, Park, Rehg, Sato, Shi, Shou, Torralba, Torresani, Yan, and Malik]{ego4d}
Kristen Grauman, Andrew Westbury, Eugene Byrne, Zachary Chavis, Antonino Furnari, Rohit Girdhar, Jackson Hamburger, Hao Jiang, Miao Liu, Xingyu Liu, Miguel Martin, Tushar Nagarajan, Ilija Radosavovic, Santhosh~Kumar Ramakrishnan, Fiona Ryan, Jayant Sharma, Michael Wray, Mengmeng Xu, Eric~Zhongcong Xu, Chen Zhao, Siddhant Bansal, Dhruv Batra, Vincent Cartillier, Sean Crane, Tien Do, Morrie Doulaty, Akshay Erapalli, Christoph Feichtenhofer, Adriano Fragomeni, Qichen Fu, Abrham Gebreselasie, Cristina Gonz\'alez, James Hillis, Xuhua Huang, Yifei Huang, Wenqi Jia, Weslie Khoo, J\'achym Kol\'a\v{r}, Satwik Kottur, Anurag Kumar, Federico Landini, Chao Li, Yanghao Li, Zhenqiang Li, Karttikeya Mangalam, Raghava Modhugu, Jonathan Munro, Tullie Murrell, Takumi Nishiyasu, Will Price, Paola Ruiz, Merey Ramazanova, Leda Sari, Kiran Somasundaram, Audrey Southerland, Yusuke Sugano, Ruijie Tao, Minh Vo, Yuchen Wang, Xindi Wu, Takuma Yagi, Ziwei Zhao, Yunyi Zhu, Pablo Arbel\'aez, David Crandall, Dima Damen, Giovanni~Maria
  Farinella, Christian Fuegen, Bernard Ghanem, Vamsi~Krishna Ithapu, C.~V. Jawahar, Hanbyul Joo, Kris Kitani, Haizhou Li, Richard Newcombe, Aude Oliva, Hyun~Soo Park, James~M. Rehg, Yoichi Sato, Jianbo Shi, Mike~Zheng Shou, Antonio Torralba, Lorenzo Torresani, Mingfei Yan, and Jitendra Malik.
\newblock Ego4d: Around the world in 3,000 hours of egocentric video.
\newblock In \emph{CVPR}, 2022.

\bibitem[Grauman et~al.(2024)Grauman, Westbury, Torresani, Kitani, Malik, Afouras, Ashutosh, Baiyya, Bansal, Boote, and et~al.]{Grauman_2024_CVPR}
Kristen Grauman, Andrew Westbury, Lorenzo Torresani, Kris Kitani, Jitendra Malik, Triantafyllos Afouras, Kumar Ashutosh, Vijay Baiyya, Siddhant Bansal, Bikram Boote, and et al.
\newblock Ego-exo4d: Understanding skilled human activity from first- and third-person perspectives.
\newblock In \emph{CVPR}, 2024.

\bibitem[Harley et~al.(2022)Harley, Fang, and Fragkiadaki]{harley2022particle}
Adam~W. Harley, Zhaoyuan Fang, and Katerina Fragkiadaki.
\newblock Particle video revisited: {T}racking through occlusions using point trajectories.
\newblock In \emph{ECCV}, 2022.

\bibitem[Hatano et~al.(2024{\natexlab{a}})Hatano, Hachiuma, Fujii, and Saito]{hatano2024eccv}
Masashi Hatano, Ryo Hachiuma, Ryo Fujii, and Hideo Saito.
\newblock Multimodal cross-domain few-shot learning for egocentric action recognition.
\newblock In \emph{ECCV}, 2024{\natexlab{a}}.

\bibitem[Hatano et~al.(2024{\natexlab{b}})Hatano, Hachiuma, and Saito]{hatano2024eccvw}
Masashi Hatano, Ryo Hachiuma, and Hideo Saito.
\newblock Emag: Ego-motion aware and generalizable 2d hand forecasting from egocentric videos.
\newblock In \emph{ECCV Workshop}, 2024{\natexlab{b}}.

\bibitem[Ho et~al.(2020)Ho, Jain, and Abbeel]{ho2020denoising}
Jonathan Ho, Ajay Jain, and Pieter Abbeel.
\newblock Denoising diffusion probabilistic models.
\newblock In \emph{NeurIPS}, 2020.

\bibitem[Hu et~al.(2024)Hu, Yin, Haeufle, Schmitt, and Bulling]{hu2024hoimotion}
Zhiming Hu, Zheming Yin, Daniel Haeufle, Syn Schmitt, and Andreas Bulling.
\newblock Hoimotion: Forecasting human motion during human-object interactions using egocentric 3d object bounding boxes.
\newblock \emph{IEEE Transactions on Visualization and Computer Graphics}, 30\penalty0 (11):\penalty0 7375–7385, 2024.

\bibitem[Jia et~al.(2022)Jia, Liu, and Rehg]{jia2022generative}
Wenqi Jia, Miao Liu, and James~M. Rehg.
\newblock Generative adversarial network for future hand segmentation from egocentric video.
\newblock In \emph{ECCV}, 2022.

\bibitem[Jiang et~al.(2024)Jiang, Streli, Meier, and Holz]{jiang2024egoposer}
Jiaxi Jiang, Paul Streli, Manuel Meier, and Christian Holz.
\newblock Egoposer: Robust real-time egocentric pose estimation from sparse and intermittent observations everywhere.
\newblock In \emph{ECCV}, 2024.

\bibitem[Karaev et~al.(2024)Karaev, Rocco, Graham, Neverova, Vedaldi, and Rupprecht]{karaev23cotracker}
Nikita Karaev, Ignacio Rocco, Benjamin Graham, Natalia Neverova, Andrea Vedaldi, and Christian Rupprecht.
\newblock Cotracker: It is better to track together.
\newblock In \emph{ECCV}, 2024.

\bibitem[Khirodkar et~al.(2023)Khirodkar, Bansal, Ma, Newcombe, Vo, and Kitani]{khirodkar2023ego}
Rawal Khirodkar, Aayush Bansal, Lingni Ma, Richard Newcombe, Minh Vo, and Kris Kitani.
\newblock Ego-humans: An ego-centric 3d multi-human benchmark.
\newblock In \emph{ICCV}, 2023.

\bibitem[Kwon et~al.(2021)Kwon, Tekin, Stuhmer, Bogo, and Pollefeys]{Kwon_2021_H2O}
Taein Kwon, Bugra Tekin, Jan Stuhmer, Federica Bogo, and Marc Pollefeys.
\newblock H2o: Two hands manipulating objects for first person interaction recognition.
\newblock In \emph{ICCV}, 2021.

\bibitem[Li et~al.(2023)Li, Liu, and Wu]{li2023ego}
Jiaman Li, Karen Liu, and Jiajun Wu.
\newblock Ego-body pose estimation via ego-head pose estimation.
\newblock In \emph{CVPR}, 2023.

\bibitem[Li et~al.(2022{\natexlab{a}})Li, Chen, Zhang, Xie, Tian, and Zhang]{li2022skeleton}
Maosen Li, Siheng Chen, Zijing Zhang, Lingxi Xie, Qi Tian, and Ya Zhang.
\newblock Skeleton-parted graph scattering networks for 3d human motion prediction.
\newblock In \emph{ECCV}, 2022{\natexlab{a}}.

\bibitem[Li et~al.(2022{\natexlab{b}})Li, Cao, Liang, Liang, Chen, Zhao, and Feng]{Li_2022_EGOPAT}
Yiming Li, Ziang Cao, Andrew Liang, Benjamin Liang, Luoyao Chen, Hang Zhao, and Chen Feng.
\newblock Egocentric prediction of action target in 3d.
\newblock In \emph{CVPR}, 2022{\natexlab{b}}.

\bibitem[Liu et~al.(2020)Liu, Tang, Li, and Rehg]{liu2020forecasting}
Miao Liu, Siyu Tang, Yin Li, and James~M Rehg.
\newblock Forecasting human-object interaction: joint prediction of motor attention and actions in first person video.
\newblock In \emph{ECCV}, 2020.

\bibitem[Liu et~al.(2022)Liu, Tripathi, Majumdar, and Wang]{liu2022joint}
Shaowei Liu, Subarna Tripathi, Somdeb Majumdar, and Xiaolong Wang.
\newblock Joint hand motion and interaction hotspots prediction from egocentric videos.
\newblock In \emph{CVPR}, 2022.

\bibitem[Luo et~al.(2021)Luo, Hachiuma, Yuan, and Kitani]{Luo2021DynamicsRegulatedKP}
Zhengyi Luo, Ryo Hachiuma, Ye Yuan, and Kris Kitani.
\newblock Dynamics-regulated kinematic policy for egocentric pose estimation.
\newblock In \emph{NeurIPS}, 2021.

\bibitem[Ma et~al.(2024{\natexlab{a}})Ma, Chen, Bao, Xu, and Wang]{ma2024madiff}
Junyi Ma, Xieyuanli Chen, Wentao Bao, Jingyi Xu, and Hesheng Wang.
\newblock Madiff: Motion-aware mamba diffusion models for hand trajectory prediction on egocentric videos.
\newblock \emph{arXiv preprint arXiv:2409.02638}, 2024{\natexlab{a}}.

\bibitem[Ma et~al.(2024{\natexlab{b}})Ma, Xu, Chen, and Wang]{ma2024diff}
Junyi Ma, Jingyi Xu, Xieyuanli Chen, and Hesheng Wang.
\newblock Diff-ip2d: Diffusion-based hand-object interaction prediction on egocentric videos.
\newblock \emph{arXiv preprint arXiv:2405.04370}, 2024{\natexlab{b}}.

\bibitem[Ma et~al.(2022)Ma, Nie, Long, Zhang, and Li]{ma2022progressively}
Tiezheng Ma, Yongwei Nie, Chengjiang Long, Qing Zhang, and Guiqing Li.
\newblock Progressively generating better initial guesses towards next stages for high-quality human motion prediction.
\newblock In \emph{CVPR}, 2022.

\bibitem[Mahmood et~al.(2019)Mahmood, Ghorbani, Troje, Pons-Moll, and Black]{mahmood2019amass}
Naureen Mahmood, Nima Ghorbani, Nikolaus~F Troje, Gerard Pons-Moll, and Michael~J Black.
\newblock Amass: Archive of motion capture as surface shapes.
\newblock In \emph{ICCV}, 2019.

\bibitem[Mao et~al.(2021)Mao, Liu, and Salzmann]{mao2021generating}
Wei Mao, Miaomiao Liu, and Mathieu Salzmann.
\newblock Generating smooth pose sequences for diverse human motion prediction.
\newblock In \emph{ICCV}, 2021.

\bibitem[M\"uller et~al.(2024)M\"uller, Ye, Pavlakos, Black, and Kanazawa]{mueller2023buddi}
Lea M\"uller, Vickie Ye, Georgios Pavlakos, Michael Black, and Angjoo Kanazawa.
\newblock Generative proxemics: A prior for 3d social interaction from images.
\newblock In \emph{CVPR}, 2024.

\bibitem[Nagarajan et~al.(2023)Nagarajan, Ramakrishnan, Desai, Hillis, and Grauman]{nagarajan2023egoenv}
Tushar Nagarajan, Santhosh~Kumar Ramakrishnan, Ruta Desai, James Hillis, and Kristen Grauman.
\newblock Egoenv: Human-centric environment representations from egocentric video.
\newblock In \emph{NeurIPS}, 2023.

\bibitem[Ng et~al.(2020)Ng, Xiang, Joo, and Grauman]{ng2020you2me}
Evonne Ng, Donglai Xiang, Hanbyul Joo, and Kristen Grauman.
\newblock You2me: Inferring body pose in egocentric video via first and second person interactions.
\newblock In \emph{CVPR}, 2020.

\bibitem[Oh et~al.(2023)Oh, Park, Kim, Moon, and Lee]{oh2023BlurHand}
Yeounguk Oh, JoonKyu Park, Jaeha Kim, Gyeongsik Moon, and Kyoung~Mu Lee.
\newblock Recovering 3d hand mesh sequence from a single blurry image: A new dataset and temporal unfolding.
\newblock In \emph{CVPR}, 2023.

\bibitem[Park et~al.(2022)Park, Oh, Moon, Choi, and Lee]{Park_2022_CVPR_HandOccNet}
JoonKyu Park, Yeonguk Oh, Gyeongsik Moon, Hongsuk Choi, and Kyoung~Mu Lee.
\newblock Handoccnet: Occlusion-robust 3d hand mesh estimation network.
\newblock In \emph{CVPR}, 2022.

\bibitem[Pavlakos et~al.(2024)Pavlakos, Shan, Radosavovic, Kanazawa, Fouhey, and Malik]{pavlakos2024reconstructing}
Georgios Pavlakos, Dandan Shan, Ilija Radosavovic, Angjoo Kanazawa, David Fouhey, and Jitendra Malik.
\newblock Reconstructing hands in 3{D} with transformers.
\newblock In \emph{CVPR}, 2024.

\bibitem[Pirsiavash and Ramanan(2012)]{hamed2012cvpr}
Hamed Pirsiavash and Deva Ramanan.
\newblock Detecting activities of daily living in first-person camera views.
\newblock In \emph{CVPR}, 2012.

\bibitem[Plizzari et~al.(2023)Plizzari, Perrett, Caputo, and Damen]{Plizzari2023iccv}
Chiara Plizzari, Toby Perrett, Barbara Caputo, and Dima Damen.
\newblock What can a cook in italy teach a mechanic in india? action recognition generalisation over scenarios and locations.
\newblock In \emph{ICCV}, 2023.

\bibitem[Poleg et~al.(2016)Poleg, Ephrat, Peleg, and Arora]{poleg2016wacv}
Yair Poleg, Ariel Ephrat, Shmuel Peleg, and Chetan Arora.
\newblock Compact cnn for indexing egocentric videos.
\newblock In \emph{WACV}, 2016.

\bibitem[Prakash et~al.(2024)Prakash, Tu, Chang, and Gupta]{prakash2024hands}
Aditya Prakash, Ruisen Tu, Matthew Chang, and Saurabh Gupta.
\newblock 3d hand pose estimation in everyday egocentric images.
\newblock In \emph{ECCV}, 2024.

\bibitem[Rempe et~al.(2021)Rempe, Birdal, Hertzmann, Yang, Sridhar, and Guibas]{rempe2021humor}
Davis Rempe, Tolga Birdal, Aaron Hertzmann, Jimei Yang, Srinath Sridhar, and Leonidas~J Guibas.
\newblock Humor: 3d human motion model for robust pose estimation.
\newblock In \emph{ICCV}, 2021.

\bibitem[Rong et~al.(2021)Rong, Shiratori, and Joo]{rong2021frankmocap}
Yu Rong, Takaaki Shiratori, and Hanbyul Joo.
\newblock Frankmocap: A monocular 3d whole-body pose estimation system via regression and integration.
\newblock In \emph{ICCV Workshop}, 2021.

\bibitem[Roy et~al.(2024)Roy, Rajendiran, and Fernando]{roy2024interaction}
Debaditya Roy, Ramanathan Rajendiran, and Basura Fernando.
\newblock Interaction region visual transformer for egocentric action anticipation.
\newblock In \emph{WACV}, 2024.

\bibitem[Sch\"oller et~al.(2020)Sch\"oller, Aravantinos, Lay, and Knoll]{cvm}
Christoph Sch\"oller, Vincent Aravantinos, Florian Lay, and Alois Knoll.
\newblock What the constant velocity model can teach us about pedestrian motion prediction.
\newblock \emph{IEEE Robotics and Automation Letters (RA-L)}, 5\penalty0 (2):\penalty0 1696--1703, 2020.

\bibitem[Sener et~al.(2022)Sener, Chatterjee, Shelepov, He, Singhania, Wang, and Yao]{sener2022assembly101}
F. Sener, D. Chatterjee, D. Shelepov, K. He, D. Singhania, R. Wang, and A. Yao.
\newblock Assembly101: A large-scale multi-view video dataset for understanding procedural activities.
\newblock In \emph{CVPR}, 2022.

\bibitem[Shan et~al.(2020)Shan, Geng, Shu, and Fouhey]{shan20_100doh}
Dandan Shan, Jiaqi Geng, Michelle Shu, and David Fouhey.
\newblock Understanding human hands in contact at internet scale.
\newblock In \emph{CVPR}, 2020.

\bibitem[Tang et~al.(2024)Tang, Zhang, Luo, Liu, and Li]{tang2025prompting}
Bowen Tang, Kaihao Zhang, Wenhan Luo, Wei Liu, and Hongdong Li.
\newblock Prompting future driven diffusion model for hand motion prediction.
\newblock In \emph{ECCV}, 2024.

\bibitem[Tekin et~al.(2019)Tekin, Bogo, and Pollefeys]{tekin2019cvpr}
Bugra Tekin, Federica Bogo, and Marc Pollefeys.
\newblock H+o: Unified egocentric recognition of 3d hand-object poses and interactions.
\newblock In \emph{CVPR}, 2019.

\bibitem[Thakur et~al.(2024)Thakur, Beyan, Morerio, Murino, and Del~Bue]{thakur2024leveraging}
Sanket Thakur, Cigdem Beyan, Pietro Morerio, Vittorio Murino, and Alessio Del~Bue.
\newblock Leveraging next-active objects for context-aware anticipation in egocentric videos.
\newblock In \emph{WACV}, 2024.

\bibitem[Vaswani et~al.(2017)Vaswani, Shazeer, Parmar, Uszkoreit, Jones, Gomez, Kaiser, and Polosukhin]{vaswani2017attention}
Ashish Vaswani, Noam Shazeer, Niki Parmar, Jakob Uszkoreit, Llion Jones, Aidan~N Gomez, \L~ukasz Kaiser, and Illia Polosukhin.
\newblock Attention is all you need.
\newblock In \emph{NeurIPS}, 2017.

\bibitem[Wang et~al.(2021)Wang, Liu, Xu, Sarkar, and Theobalt]{wang2021estimating}
Jian Wang, Lingjie Liu, Weipeng Xu, Kripasindhu Sarkar, and Christian Theobalt.
\newblock Estimating egocentric 3d human pose in global space.
\newblock In \emph{ICCV}, 2021.

\bibitem[Wang et~al.(2024)Wang, Cao, Luvizon, Liu, Sarkar, Tang, Beeler, and Theobalt]{wang2024egocentric}
Jian Wang, Zhe Cao, Diogo Luvizon, Lingjie Liu, Kripasindhu Sarkar, Danhang Tang, Thabo Beeler, and Christian Theobalt.
\newblock Egocentric whole-body motion capture with fisheyevit and diffusion-based motion refinement.
\newblock In \emph{CVPR}, 2024.

\bibitem[Wang et~al.(2023)Wang, Kwon, Rad, Pan, Chakraborty, Andrist, Bohus, Feniello, Tekin, Frujeri, Joshi, and Pollefeys]{HoloAssist2023}
Xin Wang, Taein Kwon, Mahdi Rad, Bowen Pan, Ishani Chakraborty, Sean Andrist, Dan Bohus, Ashley Feniello, Bugra Tekin, Felipe~Vieira Frujeri, Neel Joshi, and Marc Pollefeys.
\newblock Holoassist: an egocentric human interaction dataset for interactive ai assistants in the real world.
\newblock In \emph{ICCV}, 2023.

\bibitem[Wu et~al.(2020)Wu, Zhu, Wang, Yang, and Wu]{wu2020learning}
Yu Wu, Linchao Zhu, Xiaohan Wang, Yi Yang, and Fei Wu.
\newblock Learning to anticipate egocentric actions by imagination.
\newblock \emph{IEEE Transactions on Image Processing}, 30:\penalty0 1143--1152, 2020.

\bibitem[Yan et~al.(2024)Yan, Cui, Xie, and Guo]{yan2024forecasting}
Haitao Yan, Qiongjie Cui, Jiexin Xie, and Shijie Guo.
\newblock Forecasting of 3d whole-body human poses with grasping objects.
\newblock In \emph{CVPR}, 2024.

\bibitem[Yang et~al.(2022)Yang, Ma, Zuo, Wang, Gong, and Cheng]{yang20223d}
Ji Yang, Youdong Ma, Xinxin Zuo, Sen Wang, Minglun Gong, and Li Cheng.
\newblock 3d pose estimation and future motion prediction from 2d images.
\newblock \emph{Pattern Recognition}, 124:\penalty0 108439, 2022.

\bibitem[Ye et~al.(2023{\natexlab{a}})Ye, Pavlakos, Malik, and Kanazawa]{ye2023slahmr}
Vickie Ye, Georgios Pavlakos, Jitendra Malik, and Angjoo Kanazawa.
\newblock Decoupling human and camera motion from videos in the wild.
\newblock In \emph{CVPR}, 2023{\natexlab{a}}.

\bibitem[Ye et~al.(2023{\natexlab{b}})Ye, Li, Gupta, Mello, Birchfield, Song, Tulsiani, and Liu]{ye2023affordance}
Yufei Ye, Xueting Li, Abhinav Gupta, Shalini~De Mello, Stan Birchfield, Jiaming Song, Shubham Tulsiani, and Sifei Liu.
\newblock Affordance diffusion: Synthesizing hand-object interactions.
\newblock In \emph{CVPR}, 2023{\natexlab{b}}.

\bibitem[Yi et~al.(2024)Yi, Ye, Zheng, M\"uller, Pavlakos, Ma, Malik, and Kanazawa]{yi2024egoallo}
Brent Yi, Vickie Ye, Maya Zheng, Lea M\"uller, Georgios Pavlakos, Yi Ma, Jitendra Malik, and Angjoo Kanazawa.
\newblock Estimating body and hand motion in an ego-sensed world.
\newblock \emph{arXiv preprint arXiv:2410.03665}, 2024.

\bibitem[Yuan and Kitani(2018)]{yuan20183d}
Ye Yuan and Kris Kitani.
\newblock 3d ego-pose estimation via imitation learning.
\newblock In \emph{ECCV}, 2018.

\bibitem[Yuan and Kitani(2019)]{yuan2019ego}
Ye Yuan and Kris Kitani.
\newblock Ego-pose estimation and forecasting as real-time pd control.
\newblock In \emph{ICCV}, 2019.

\bibitem[Zhang et~al.(2022{\natexlab{a}})Zhang, Zhou, Stent, and Shi]{zhang2022fine}
Lingzhi Zhang, Shenghao Zhou, Simon Stent, and Jianbo Shi.
\newblock Fine-grained egocentric hand-object segmentation: Dataset, model, and applications.
\newblock In \emph{ECCV}, 2022{\natexlab{a}}.

\bibitem[Zhang et~al.(2022{\natexlab{b}})Zhang, Ma, Zhang, Qian, Kwon, Pollefeys, Bogo, and Tang]{Zhang2022EgoBody}
Siwei Zhang, Qianli Ma, Yan Zhang, Zhiyin Qian, Taein Kwon, Marc Pollefeys, Federica Bogo, and Siyu Tang.
\newblock Egobody: Human body shape and motion of interacting people from head-mounted devices.
\newblock In \emph{ECCV}, 2022{\natexlab{b}}.

\bibitem[Zhang et~al.(2023)Zhang, Ma, Zhang, Aliakbarian, Cosker, and Tang]{zhang2023probabilistic}
Siwei Zhang, Qianli Ma, Yan Zhang, Sadegh Aliakbarian, Darren Cosker, and Siyu Tang.
\newblock Probabilistic human mesh recovery in 3d scenes from egocentric views.
\newblock In \emph{ICCV}, 2023.

\bibitem[Zheng et~al.(2023)Zheng, Liu, Qi, and Chen]{zheng2023potter}
Ce Zheng, Xianpeng Liu, Guo-Jun Qi, and Chen Chen.
\newblock Potter: Pooling attention transformer for efficient human mesh recovery.
\newblock In \emph{CVPR}, 2023.

\bibitem[Zhou et~al.(2019)Zhou, Barnes, Lu, Yang, and Li]{zhou2019continuity}
Yi Zhou, Connelly Barnes, Jingwan Lu, Jimei Yang, and Hao Li.
\newblock On the continuity of rotation representations in neural networks.
\newblock In \emph{CVPR}, 2019.

\end{thebibliography}
}

\clearpage
\setcounter{page}{1}
\maketitlesupplementary

\thispagestyle{empty}
\appendix

\section{Additional Implementation Details}
We use the ViT-Small as the visual encoder, so the dimension of image features $d_{\text{img}}$ is $384$.
As for the dimension of tokens for each timestep $d_{\text{z}}$, we set it to $512$, and the number of layers and the number of heads of the Transformer are set to $4$ and $8$.

\section{Additional Evaluation Results}
\label{sec:add-eval}
\noindent \textbf{Balancing Hyperparameters}.
We conduct an ablation study to analyze the impact of balancing hyperparameters for the reprojection loss $\mathcal{L}_{\text{reproj}}$ and the visibility loss $\mathcal{L}_{\text{vis}}$ on the hand forecasting accuracy.
We systematically vary the values of each balancing hyperparameter across predefined ranges: $\lambda_{\text{reproj}}$ from $0.01$ to $0.5$, and $\lambda_{\text{vis}}$ from $0.01$ to $1.0$.
We vary one hyperparameter at a time while keeping the others fixed at their default values: $0.05$ and $0.1$ for $\mathcal{L}_{\text{reproj}}$ and $\mathcal{L}_{\text{vis}}$, respectively.

We report the hand trajectory and pose forecasting accuracy of the proposed model with varied hyperparameters in terms of ADE and MPJPE in \cref{table:hypara}.
This ablation analysis reveals the importance of balancing these hyperparameters for optimal hand forecasting accuracy.
Specifically, a lower value of $\lambda_{\text{reproj}}$ reduces the model's reliance on consistency between 2D input and 3D output, leading to poor spatial alignment with visible 2D hand input. 
Conversely, a high value of $\lambda_{\text{reproj}}$ overemphasizes reprojection accuracy, causing the model to neglect the correct 3D depth (i.e. distance from camera) estimation and resulting in suboptimal predictions.
Regarding the weight for visibility loss, a higher value degrades the forecasting performance as it is not directly related to hand forecasting, while a lower value reduces the model's in- or out-of-view awareness, leading to a performance drop.

\begin{table}%
\captionsetup[subfloat]{position=top}
\centering
\caption{\textbf{Balancing Hyperparameters}. We conduct an ablation study on the loss weights for the reprojection loss and visibility loss, and report the hand trajectory and pose forecasting accuracy in terms of ADE and MPJPE, respectively.}
\label{table:hypara}
\subfloat[$\lambda_{\text{reproj}}$]{
\label{Tab_a}
\scalebox{1}{
    \begin{tabular}{lcc}
      \toprule 
      $\lambda_{\text{reproj}}$ & ADE & MPJPE\\
      \midrule
      $0.5$ & 0.262 & 0.125\\
      \rowcolor{yellow2}
      $0.05$ & \textbf{0.261} & \textbf{0.115}\\
      $0.01$ & 0.262 & 0.123\\
      \bottomrule
    \end{tabular}
}}%
\subfloat[$\lambda_{\text{vis}}$]{
\label{Tab_b}
\scalebox{1}{
    \begin{tabular}{lcc}
      \toprule 
      $\lambda_{\text{vis}}$ & ADE & MPJPE\\
      \midrule
      $1.0$ & 0.275 & 0.121\\
      \rowcolor{yellow2}
      $0.1$ & \textbf{0.261} & \textbf{0.115}\\
      $0.01$ & 0.273 & 0.124\\
      \bottomrule
    \end{tabular}
}}%
\end{table}

\begin{figure}
  \begin{subfigure}{0.23\textwidth}
    \includegraphics[width=\linewidth]{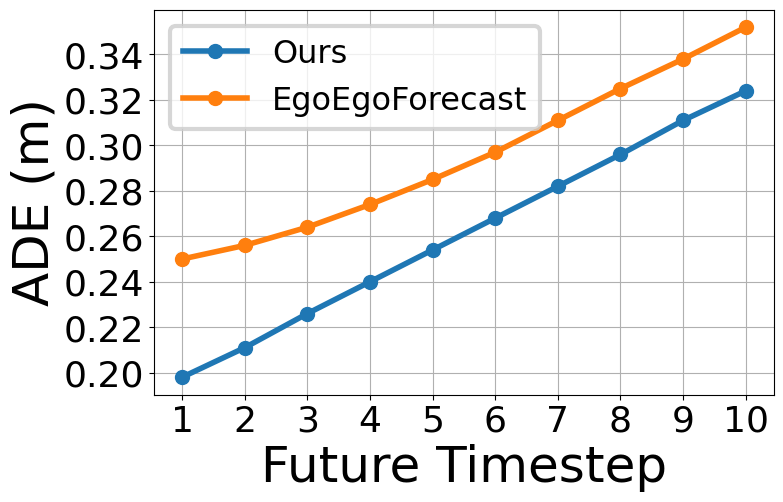}
    \caption{Per-timestep hand trajectory forecasting accuracy.}
    \label{fig:figure1}
  \end{subfigure}%
  \hfill
  \begin{subfigure}{0.23\textwidth}
    \includegraphics[width=\linewidth]{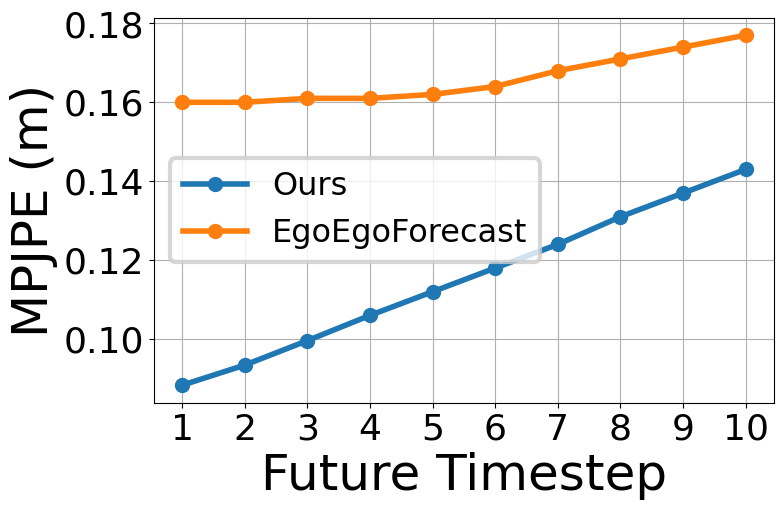}
    \caption{Per-timestep hand pose forecasting accuracy.}
    \label{fig:figure2}
  \end{subfigure}%
  \caption{\textbf{Per-timestep Hand Forecasting Accuracy}. We report the hand trajectory forecasting accuracy in ADE and hand pose forecasting accuracy in MPJPE for every future timestep. Lines in \textcolor{blue}{blue} and \textcolor{BurntOrange}{orange} represent the performance of our model and the EgoEgoForecast baseline, respectively.}
  \label{fig:per-timestep}
\end{figure}

\noindent \textbf{Per-timestep Hand Forecasting Accuracy}. We report the hand trajectory and pose forecasting accuracy for each future timestep in \cref{fig:per-timestep}.
Overall, \method outperforms the EgoEgoForecast baseline on every future timestep for both hand trajectory and pose forecasting tasks.
Specifically, the improvements over the baseline are most pronounced at earlier future timesteps in the hand pose forecasting, as \method achieves more accurate hand pose estimation by leveraging visible 2D hand locations. 
In the hand trajectory forecasting task, our model consistently outperforms the baseline by effectively accounting for in-view or out-of-view during the observation period.

\end{document}